\pdfoutput=1

\documentclass[11pt]{article}

\usepackage[]{acl}

\usepackage{times}
\usepackage{latexsym}

\usepackage[T1]{fontenc}

\usepackage[utf8]{inputenc}

\usepackage{microtype}

\usepackage{times}
\usepackage{latexsym}
\usepackage{hyperref}
\usepackage{graphicx}
\usepackage{graphics}
\usepackage{multirow}
\usepackage{amsmath}
\usepackage{amsfonts}
\usepackage{amssymb}
\usepackage{tablefootnote}
\usepackage{color}
\usepackage{bbm}
\usepackage{graphicx}
\usepackage{xspace}
\usepackage{tikz}
\usepackage{pgfplots}
\usepackage[normalem]{ulem}
\usepackage{xcolor}
\usepackage{dashbox}
\usepackage{textcomp}
\usepackage{tcolorbox}
\usepackage{adjustbox}
\usepackage{url}
\usepackage{subcaption}
\usepackage{fdsymbol}
\usepackage{adjustbox}
\usepackage{tikz}
\usepackage{pgfplots}
\usepackage[scientific-notation=true]{siunitx}
\usepackage{booktabs}
\usepackage{wrapfig}
\definecolor{lightblue}{HTML}{E0ECF7}
\definecolor{darkblue}{HTML}{092E6B}
\newcommand{\win}[1]{{\colorbox{lightblue}{ #1}}}
\newcommand{\lose}[1]{{\colorbox{darkblue}{\color{white}{#1}}}}
\title{Discourse-Aware Soft Prompting for Text Generation}

\author{
 Marjan Ghazvininejad \ \ \ \ \ \ \ \ \ \ 
 Vladimir Karpukhin \ \ \ \ \ \ \ \ \ \ \
 Vera Gor\ \ \ \ \ \ \ \ \ \ \
 Asli Celikyilmaz \ \ \ \ \ \ \ \ \ \\
 ~~Facebook AI Research\\
 \small\texttt{ \{ghazvini, vladk, vgor, aslic\}@fb.com}\\
}

\begin{document}
\maketitle
\begin{abstract}
Current efficient fine-tuning methods
(e.g., adapters \cite{adapters}, prefix-tuning \cite{prefixtune}, etc.) 
have optimized conditional text generation via training a small set of extra parameters of the neural language model, while freezing the rest for efficiency. While showing strong performance on some generation tasks, they don't generalize across all generation tasks. We show that soft-prompt based conditional text generation can be improved with simple and efficient methods that simulate modeling the discourse structure of human written text.
We investigate two design choices: 
First, we apply \textit{hierarchical blocking} on the prefix parameters to simulate a higher-level discourse structure of human written text. Second, we apply \textit{attention sparsity} on the prefix parameters at different layers of the network and learn sparse transformations on the softmax-function. We show that structured design of prefix parameters yields more coherent, faithful and relevant generations than the baseline prefix-tuning on all generation tasks.\footnote{All supporting code will be publicly released.} 

\end{abstract}

\section{Introduction}

Recent advances in pre-trained langauge models (PLMs) \citep{lewis-etal-2020-bart, 2020t5, radford2019language} have made great impact on text generation research, especially when they are fine-tuned on downstream tasks such as summarization, data-to-text generation, long-question answering, etc. Consequent research have shown that PLMs' impact can further be improved when trained with more parameters, on more data and with more compute (GPT-3 \citep{gpt3}, Megatron \citep{megatron}). On the flip side, storing larger LMs or fully fine-tuning them (updating all the parameters) on downstream tasks usually causes resource or over-fitting issues.

To mitigate fine-tuning issues, recent work have proposed \textit{prompt-based learning} \citep{prompttuning_survey}, which focus on learning textual prompts to steer PLMs' continuation towards desired output while keeping the model parameters frozen. While providing strong control of the PLMs, such prompt engineering could be time consuming requiring manual crafting.
There is a growing research direction under prompt learning towards lightweight fine-tuning \citep{adapters,Lester2021ThePO}, which update only a small number of existing or extra parameters while keeping most of the original pre-trained parameters frozen. Among them is \textit{prefix-tuning} \citep{prefixtune}, which prepends tunable continuous task-specific prompt vectors called \textit{prefix}es to the input and only trains these continuous prompts during fine-tuning. Although prefix-tuning can yield comparable results to full fine-tuning on some generation tasks, it did not generalize to tasks like abstractive summarization. 

In this work, we focus on prefix-tuning and investigate ways to improve its generalization on text generation tasks. We start asking the following questions that motivates our design choices:
(1) \textit{Do different parts of the transformer network process the prefix parameters more efficiently?};
(2) \textit{Do prefix parameters capture high-level discourse structure of the input text?}; 
(3) \textit{Can constraining prefix attention distribution to be structurally sparse enable better transfer of the task features?}

To address (1), we conduct empirical analysis on prefix-tuned BART \citep{lewis-etal-2020-bart}, by varying the size of the prefix parameters at the encoder and decoder networks on text generation tasks. 
We find that the prefix parameters at higher layers impact the performance the most, while sparse prefixes can be sufficient at the lower layers ($\S$~\ref{exp:prefixanalysis}).  

Motivated by this finding and to address (2), we investigate \textbf{discourse-aware soft prompting} via \textbf{hierarchical blocking} of prefix parameters. Previous text generation work (e.g., abstractive summarization) has shown that abstraction can be better modeled with hierarchically structured architectures \cite{liu2019,fabbri-etal-2019-multi,xiao2021primer}. To simulate a hierarchical discourse structure while \textit{only} tuning additional prefix parameters, we first split the input and output text into segments  and then assign sets of prefix parameters to each segment at different layers. 
With this structure, a set of prefixes can only be reached by their designated input or output segments during self-attention. We argue that for conditional generation tasks with hierarchically structured blocking of prefixes, we can simulate the structure of human writing styles: in input text each paragraph is a distinct section of related sentences and in output text (e.g., summary) each output sentence outlines salient concepts. Thus, a set of prefixes designated to each input and output segment at different layers can learn levels of abstractions from each section. We show performance improvements over baseline prefix tuning, yielding comparable results to full fine-tuning in several generation tasks in $\S$~\ref{blockingresults}.  

Inspired by these findings, we address (3) by applying a suite of known \textbf{sparse attention} alternatives to standard full-attention matrix during prefix-tuning. Our goal is to analyze whether sparse prefix-tuned models can encode important features better than dense prefix-tuned models. Prior work have shown that sparsity in self-attention not only improves training efficiency, but also focusing on \textit{salient features} while pushing down unrelated features and relations can impact the model performance. This improves language modeling \citep{expirespan,linformer}, language understanding \citep{Shi2021SparseBERTRT, cui-etal-2019-fine} and text generation \citep{bigbird,Li2021EASEES,liu2021hetformer,manakul-gales-2021-long}. Motivated by previous work we \textbf{inject sparsity into the self-attention} by substituting the softmax function with a 
sparse alternative under encoder prefix-tuning (without introducing any additional model parameters). 
With spectral analysis we show that sparse prefix parameters can identify important features compared to dense prefix parameters. Our quantitative analysis yield performance improvements on automatic metrics over best prefix-tuning models, while human judges generally prefer our sparse prefix model generations on factuality and coherence criteria ($\S$~\ref{doessparseattentionlhelp} and $\S$~\ref{sparsityinhierblock}). 
 
 Efficient tuning of PLMs offers a promising new direction for many NLP tasks including text generation, which we study in this work. 
 Our results suggest that \textbf{prompt design with hierarchical structure and sparsity in prefix parameters}: (\textit{i}) generate more coherent and faithful text than baseline prefix-tuning across several summarization and data-to-text generation tasks, (\textit{ii}) trail the performance of fine-tuning on most summarization tasks with a small margin, while at par with fine-tuning on data-to-text generation tasks, (\textit{iii}) outperform all the baselines in low-resource settings.

\section{Related Work}
\textbf{Prompt Tuning.} Recent years have observed several efficient methods for prompt-based tuning of large-scale PLMs \citep{prompttuning_survey}. These range from prompt engineering \citep{petroni-etal-2019-language,cui-etal-2021-template}, 
to more advanced approaches such as prompt ensembling \citep{Mao2021UniPELTAU}, composition \citep{hanpromptcomp}, or prompt-aware training methods \citep{Lester2021ThePO,promptdesign}. \citet{prefixtune} propose \textit{prefix-tuning} and show strong results on some text generation tasks, leaving room for further generalization. Here, we build directly upon
the prefix-tuning from \citet{prefixtune}, showing
where it falls short and providing several discourse-aware prompt design approaches. 
We find with human evaluations ($\S$~\ref{humanevals}) on relevance criteria that the
prefix-tuning struggles with encoding of salient concepts that constraint 
generation models require. This setting bears similarities to 
discourse modeling, which we discuss below.

\noindent\textbf{Discourse Modeling.} 
A large family
of methods make architectural design choices 
to teach models about the overall document discourse structure
\citep{marcu-1997-discourse,Barzilay2004CatchingTD, coherence2, coherence3} to improve the summarization task.
Recent work investigate different architectures to model the discourse structure via: structured attention \citep{cohan-etal-2018-discourse}, graph based methods \citep{dong-etal-2021-discourse}, or hierarchical encoders \citep{pasunuru2021data}. We simulate the discourse structure of text via hierarchical prefix structure and propose discourse-aware prompt-design for efficient PLM tuning.

\noindent\textbf{Sparse Language Models.} 
Most work on sparsity in transformers aim at improving the time and space 
bottleneck of dense transformers \citep{tay2021long}. 
Work on text generation imbue sparsity to
improve coherence, fluency, \textit{n}-gram diversity and reduce repetition. These work range from: sparse methods on
posterior vocabulary distributions at inference time \citep{Fan2018HierarchicalNS,nucleus}, sparse attention mechanisms \citep{cui-etal-2019-fine,liu2021hetformer,Shi2021SparseBERTRT, expirespan}, 
modified softmax \citet{Martins2020SparseTG}, or loss functions \citep{unlikelihood} to improve LM coherence and generalization. Similarly, we inject sparsity on the attention matrix of prefix+input features to improve the knowledge transferred to downstream text generation tasks and generating more relevant and coherent text ($\S$\ref{experimentresults}).

\section{Prefix-Tuning}
\begin{figure}[t]
    \centering
    \adjustbox{trim={.01\width} {.01\height} {0.01\width} {.01\height},clip}
    {\includegraphics[scale=0.30]{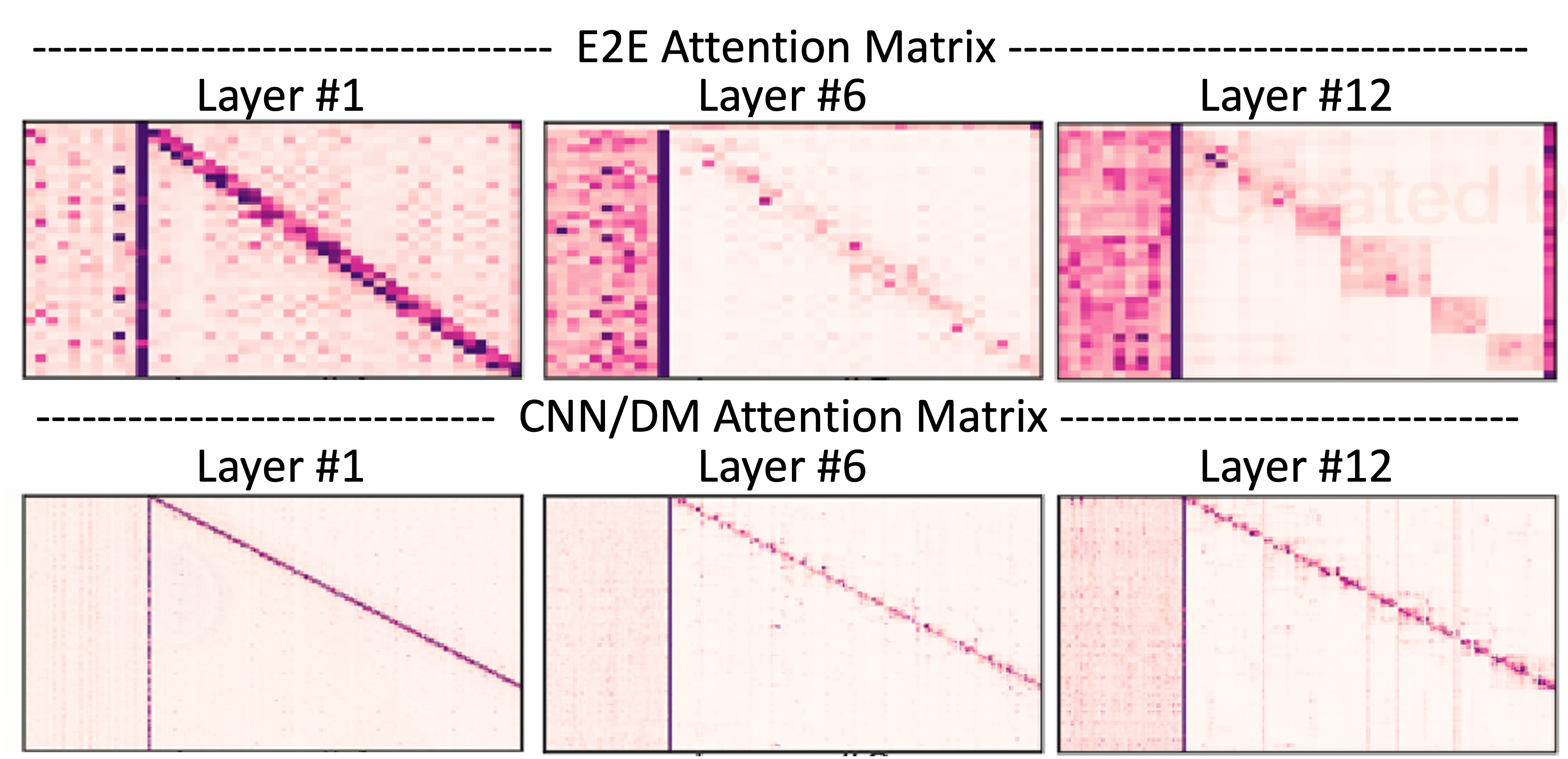} }
    \caption{\small Encoder self-attention matrices \textbf{\textit{A}} from layers 1, 6 and 12 of prefix-tuned models showing query attention scores (on \textit{y}-axis) over all prefix+inputs keys (on \textit{x}-axis). 
    Top row are matrices of models on E2E dataset where the first 10 features on \textit{x}-axis are prefix features, and bottom row are on CNN/DM dataset where first 100 features are prefix parameters.}%
    \label{fig:attn}%
\end{figure}


Following the intuition of the text-based prompt tuning methods \citep{prompttuning_survey}, prefix-tuning \citep{prefixtune} introduces task-specific prompt parameters with the goal of triggering the desired response of the LM without updating any of the original LM parameters.
At each layer, it prepends tunable prefix parameters (also called \textit{soft-prompts}) as additional keys and values to the multi-head self-attention. 
Prefix-tuning defines $h_i^{(l)}$ as the activation at the \textit{i}-th token ($i$=1$\cdots$T) of the \textit{l}-th layer in a \textit{L}-layer transformer:   
\begin{equation}
h_i^{(l)} =
    \begin{cases}
      P_{\theta}[i,:], & \text{if} i \in P_{\text{idx}}\\
      LM_{\phi}(z_i,h_{<i}) & $otherwise$\\
    \end{cases}  
\end{equation}
$[,]$ indicates concatenation, $P_{\text{idx}}$ is the sequence of prefix indices and where the activations of the first $|P_{\text{idx}}|$ positions are directly calculated by $P_\theta$ and $z_i$ is the \textit{i}-th token in the input sequence. 
During training only the parameters corresponding to the prefix keys and values are updated and the same objective function as finetuning is used\footnote{For details on prefix-tuning, pls. see \citep{prefixtune}.}.

\section{Discourse Aware Prefix-Tuning}
\label{discourse}
\noindent\textbf{Visualizing the prompt impact.} 
To motivate the discourse-aware prompt design, we investigate the impact of prefix-parameters on transformer models during prefix-tuning. 
We first analyze the attention behaviour similar to \citep{sun-lu-2020-understanding}. We prefix-tune two \textsc{Bart-Large} models, one on data-to-text generation task with E2E dataset \citep{e2e}, and another on summarization with CNN/DM \citep{hermann2015teaching}. For E2E we use 10-prefixes (the first 10 keys are from prefix parameters) and 100-prefixes for CNN/DM\footnote{The length of per instance prefix+input tokens is 100+512 in CNN/DM and 10+16 in E2E dataset.} (similar to \citet{prefixtune}). In Figure~\ref{fig:attn}, we plot the encoder self-attention distributions \textbf{\textit{A}} for different layers averaging over all head vectors. The \textit{x}-axis represents the keys while y-axis denotes the queries. For attention matrices of all the layers, see Appendix \ref{app:prefixvisual} Figure~\ref{app:fig:attn}. The attention scores show stronger relations with the prefix-keys in the E2E model compared to CNN/DM, where the prefixes exhibit weaker relations compared to the 
input keys. We attribute this to a few issues which we investigate in this work: 

\noindent\textbf{Modeling hierarchical structure.} Firstly, during prefix-tuning, the model should not only focus on learning the task specific semantics, but also the models should learn the corresponding discourse structure of the downstream task datasets. To model the intrinsic structure of the input text, biasing transformer models with a type of hierarchy has been shown to improve the generation performance. For example, previous work \citep{cohan-etal-2018-discourse,liu2019} learns the discourse structure of human written text (e.g., the beginning, body, conclusion paragraphs, topic shifts, etc.) with hierarchically structured transformers to capture the salient aspects in the input text necessary for improved performance in summarization.
With probing experiments \citet{jawahar-etal-2019-bert} show that BERT \cite{bert}
captures surface features and phrase-level information in the lower layers, 
syntactic features in the middle and semantic features and long-distance dependencies at the higher layers. Motivated by these, we introduce variations of \textbf{hierarchical blocking} on prefix parameters at different layers of the network and investigate their impact on text generation with qualitative and quantitative experiments.

\noindent\textbf{Introducing sparsity.} Secondly, the weaker prefix attention in longer inputs (Figure~\ref{fig:attn}-CNN/DM attention matrices) may imply that the attention neglects important connections, and potentially disturbed by many unrelated words. 
This issue can be attributed to the softmax function at attention score calculation \citep{Laha2018OnCS,cui-etal-2019-fine}. Softmax produces attention distribution with dense dependencies between words, and fails to assign near/exactly zero probability to less meaningful relations.
Thus, the model neglects to put more attention to important connections while also being easily disturbed by many unrelated words \citep{cui-etal-2019-fine}. This issue is more pronounced in tasks like abstractive summarization, since only a handful of salient input aspects is needed to compose a coherent summary. Sparse attention mechanisms \cite{liu2021hetformer,Shi2021SparseBERTRT} can remedy this issue by learning to put more emphasis on the important features.  

Below we describe ways to apply a suite of blocking schemes and sparsity in prefix-tuning models as sketched in Figure~\ref{fig:sparseattn}. Each block represents attention matrix  $\mathbf{A}$$\in$$\textstyle \mathbb{R}^{T \times (P+T)}$, while each row vector $a_t$$\in \mathbb{R}^{(P+T)}$, $t=1 \cdots T$, are the attention weights of $P$-prefix and $T$-input-key features.


\subsection{Prefix Blocking}
\label{prefixblocking}
\begin{figure}[t]
    \centering
    \adjustbox{trim={0\width} {.02\height} {0\width} {0\height},clip}
    {\includegraphics[width=0.48\textwidth]{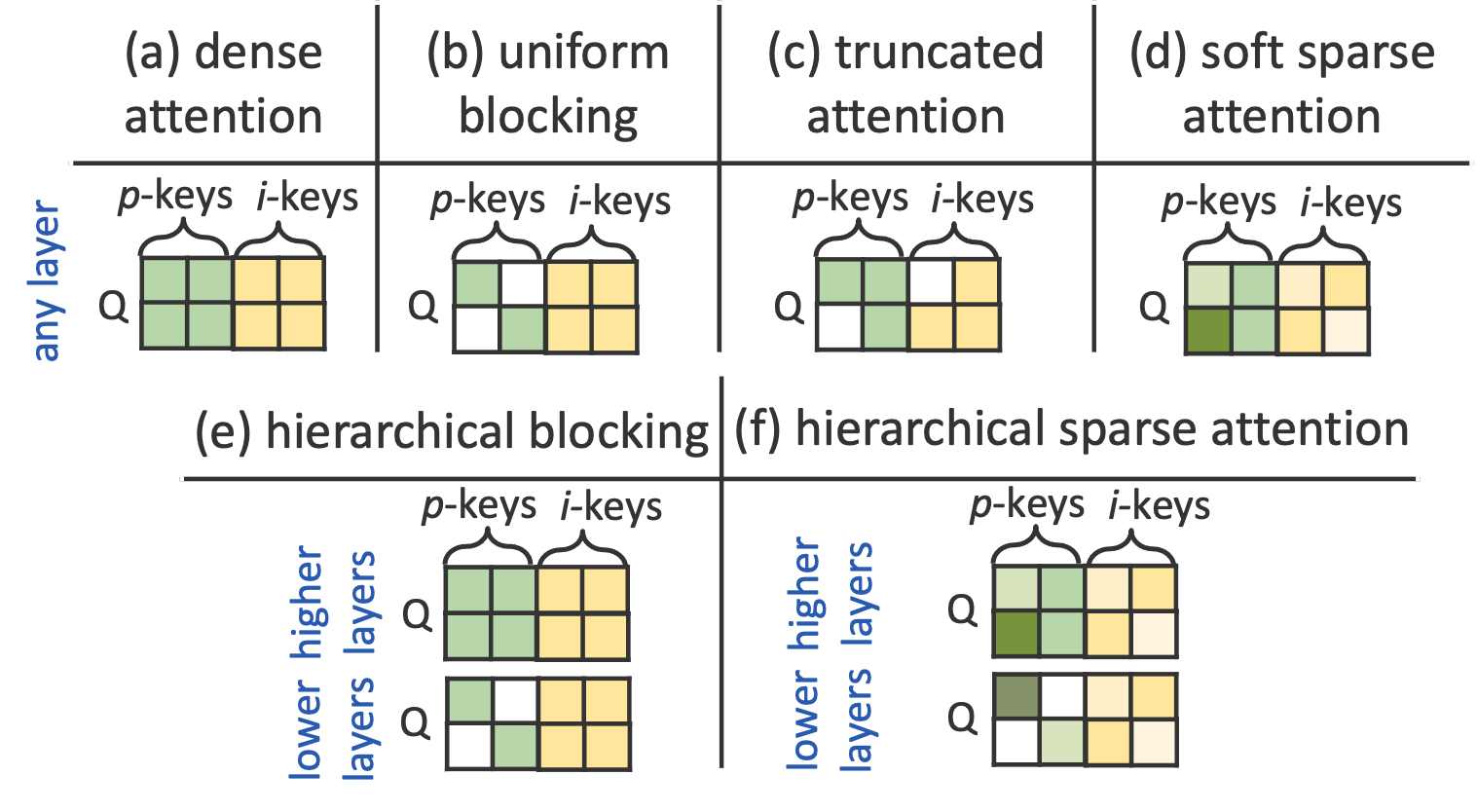} }
    \vspace{-0.5cm}
    \caption{\small Sketches of attention matrices \textbf{\textit{A}} used in prefix-tuning models representing different prefix design patterns.
    $p$-keys and $i$-keys denote $P$ prefix and $T$ input keys. 
    Sparsity of attention scores are indicated by color gradations. White cells in any row represent blocked parameters for the query.}%
     \vspace{-0.3cm}
    \label{fig:sparseattn}%
\end{figure}

As shown in Figure~\ref{fig:sparseattn}-(b) and (e), the
two variations of prefix-blocking we apply here are a type of structural bias we imbue the models to simulate high-level discourse structure of documents: 

\noindent\textbf{(\emph{i}) Uniform Blocking (UniBlock)}: 
 We first split the sequence of input tokens into segments. We allocate different sets of prefix parameters to each segment and apply blocking on the rest of the prefix parameters.   
In baseline prefix-tuning, a query of a token can bind with all the prefix and input key and value parameters, while in the uniform blocked prefix-tuning, the query of a token in the input or output segment can bind with all input key and values but only with the designated prefix key and value vectors. 
For example, if 100 prefix parameters are used and we split the input tokens into 2 segments, the first 50 prefix keys and values can only be bound with the query vectors of input tokens from the first input segment and so on. 
We only apply blocking to the prefix parameters and let all inputs tokens  attend to each other, see Figure~\ref{fig:sparseattn}-(b).   
In uniform blocking, we use the same blocking schema at each layer. 

\noindent\textbf{(\emph{ii}) Hierarchical Blocking (HierBlock)}: 
To bias the prefix parameters with a form of hierarchy, we use the uniform prefix-blocking on the lower layers of the transformer, while we let all tokens attend to all prefixes at the top layers  as shown in Figure~\ref{fig:sparseattn}-(e).
The attention matrix of the top layers is same as the standard prefix-tuning of \citep{prefixtune} where no blocking on prefixes is applied.

\subsection{Sparse Attention Prefix-Tuning}
\label{sparseattnpt}
To train a prefix-tuning model that learns to highlight important input content, we apply five sparse attention design options on the encoder. 

\noindent\textbf{(\emph{a}) Truncated Sparse Attention (TruncSA)}:
\citet{dai-etal-2021-multimodal} used
 sparse cross-attention using truncation to improve salient feature extraction which showed improvements in downstream tasks performance. To simulate encoding with salient features, we apply top-\textit{p} truncation on both the prefix and input keys as follows: we first add all the row elements $a_{ti}$ $\in $[0,1] of the attention matrix, namely the attention scores contributing from all the queries, then normalize across all key-features, which yields new key-feature row vectors $\Tilde{a}_t$ $\in \mathbb{R}^{(P+T)}$, $\Tilde{a}_t$ $\in \Tilde{A}$:
\begin{equation}
    \bar{a}_t = \textstyle \sum_i^{T} a_{ti} \quad \Tilde{a}_t = \bar{a}_t/(\sum_t^{(P+T)} \bar{a}_t)
\end{equation}
Using top-\textit{p} truncation \citep{dai-etal-2021-multimodal} we truncate the feature key scores and use 
the top-\textit{p} portion of the probably mass in each key attention score.  
We create a binary mask for each key feature via $mask(\Tilde{\mathbf{A}}) = \text{top-}p({\Tilde{\mathbf{A}}}, \tau)$ by assigning 1.0 to the keys that the top-\textit{p} sampling has selected, 0 otherwise and threshold parameter $\tau$ controls sparsity.
Lastly, we broadcast point-wise multiplication between the sparse mask and the attention matrix  $\mathbf{A}$ to obtain the top-\textit{p} sparse attention matrix $\Tilde{\mathbf{A}} = mask(\mathbf{\Tilde{A}}) \odot \mathbf{A}$, as sketched in Figure~\ref{fig:sparseattn}-(c). 

The top-\textit{p} truncation is similar to using dropout on the features of the network while controlling the dropout rate with a user-defined threshold to compensate for overfitting and performance. 
Although top-\textit{p} sparse attention provides automatic control over attention sparsity, truncation completely masks some features. 
Next, we show how to dynamically learn to apply \textit{soft}-sparsity via sampling from a distribution. 

\par \noindent\textbf{(\emph{b}) \textit{Soft} Sparse Attention (SoftSA)}:
Influencing the attention distribution with a stochastic mask to attend to salient tokens can potentially help build higher quality sparse attention for text modeling.
Several work investigate novel approaches to learn the sparsity in attention matrix \cite{Li2021EASEES,10.1162/tacl_a_00353,Shi2021SparseBERTRT} using a sampling method to formulate the right amount of sparsity. They associate the attention scores $a_{ti}$ with each cell ($t,i$) in $\mathbf{A}$ and define a sampling distribution to learn the attention mask during training as sketched in Figure~\ref{fig:sparseattn}-(d). Similarly, we apply relaxed Bernoulli distribution as a sampler to construct our stochastic mask. 
Since sampling from Bernoulli distribution is not differentible, we use the
Gumbel Softmax reparameterization trick \citep{jang2017categorical}
with gumbel-softmax:
\begin{equation}
\tilde{a}_{t} = \underset{n \in 1\dots (P+T)}{\text{Softmax}} \left(a_{tn},g,\tau\right)
\end{equation}
\noindent where $g$=$-\log(-\log(u))$ is an independent Gumbel noise generated
from the uniform distribution $u \sim U(0, 1)$
and $\tau$ is a temperature. 
As $\tau$ approaches zero, the gumbel output approaches to a discrete distribution in \{0, 1\}, becomes identical to
those from the Bernoulli distribution. For details on Gumbel-softmax, see Appendix~\ref{app:hyperparameters}.

\noindent\textbf{(\emph{c \& d}) Hierarchical Sparse Attention}:
To simulate an intrinsic discourse structure of the input text, similar to the hierarchical blocking in $\S$~\ref{prefixblocking}, we apply sparsity on the parameters only at the lower layers. We train hierarchical models with the dense attention at the higher layers, and apply (\textit{c}) \textit{truncated} (\textbf{HTruncSA})  or (\textit{d}) \textit{soft} sparse attention (\textbf{HSoftSA}) at the lower layers (see Figure~\ref{fig:sparseattn}-(f)).
\par \noindent\textbf{(\emph{e}) Hierarchical Blocking with Sparse Attention (HierBlock+SoftSA)}:
The hierarchical blocking models we used in $\S$~\ref{prefixblocking} puts restrictions on the prefix parameters that input tokens can bind with at different layers of the network. To analyze the impact of\textbf{ ensemble of prefix blocking and sparsity}, we apply sparsity on the hierarchically blocked prefix-tuning models. 
We apply soft sparsity (SoftSA) on the lower layers of the network attention matrices of HierBlock models and keep the 
higher layer attention matrices dense.

\begin{table}[t]
    \centering
    \small
    \begin{tabular}{l l c}
        \hline 
        Dataset & Domain & \#Data  \\
         \hline
         \textbf{Summarization} & & Train/Val/Test \\
         \hline
         XSum (\citeyear{narayan-etal-2018-dont}) &  News &  204K/11K/11K  \\
         CNN/DM (\citeyear{hermann2015teaching})   &  News &  287K/13K/11K \\
         Wikihow (\citeyear{koupaee2018wikihow}) & DIY & 157K/5.6K/5.6k   \\
         SAMSum (\citeyear{gliwa2019samsum})   &  Dialog &   15.7K/<1K/<1K \\
         Pubmed (\citeyear{cohan-etal-2018-discourse})  & Clinical & 203K/6K/6K   \\
         \hline
         \textbf{Structure to Text (S2T)} & & \\
         \hline
         E2E (\citeyear{novikova2017e2e,e2e})  & Reviews & 33K/4K/4.7K   \\
         DART (\citeyear{nan2021dart}) & Reviews& 63K/7K/12.5K\\
        \hline
    \end{tabular}
    \vspace{-8pt}
    \caption{\small Datasets used in the experiments.}
    \label{tab:datasets}
    \vspace{-0.3cm}
\end{table}

\section{Experiment Setup}
\label{setup}
\noindent{\textbf{Methods.}} 
We build all fine/prefix-tuning models on the multi-layered encoder-decoder Transformer architecture using \textsc{Bart-Large} \citep{lewis-etal-2020-bart}, though our methods can be applied to any transformer architecture with key-value attention.
We compare our discourse aware prefix-tuning approaches to full parameter fine-tuning and baseline prefix-tuning \citep{prefixtune}. Finetuning updates all the LM parameters, while all prefix-tuning models freeze LM parameters and only update the prefix parameters. Baseline prefix-tuning models update prefix parameters at each layer (full-stack) of the transformers using dense attention while our proposed models use variations of sparse and blocked attention at different layers of the network.  
We choose the best models on validation dataset during training.
For details of the setup see Appendix~\ref{app:hyperparameters}.

\begin{table*}[t] \fontsize{8}{9}\selectfont
    \centering
    
    \begin{tabular}{@{}lll@{}l@{}l@{}l@{}}
        \hline 
        & \multicolumn{1}{c}{\textbf{XSum}}   &  \multicolumn{1}{c}{\textbf{CNN/DM}}  \ \ \ \ \ \  & \multicolumn{1}{c}{\textbf{PubMed}}  \ \ \ \ & \multicolumn{1}{c}{\textbf{Wikihow}} \ \ \ \  & \multicolumn{1}{c}{\textbf{SAMSum}}\\
        \textbf{Method} & \multicolumn{1}{c}{\textbf{R1/R2/R-L}} &
        \multicolumn{1}{c}{\textbf{R1/R2/R-L}} & \multicolumn{1}{c}{\textbf{R1/R2/R-L}} & \multicolumn{1}{c}{\textbf{R1/R2/R-L}} & \multicolumn{1}{c}{\textbf{R1/R2/R-L}} \\
         \hline
         \multicolumn{6}{l}{{\textbf{Fineune (FT)}}} \\
         \hline
          FT \textbf{[*]}    & 
          45.14/22.27/37.25 $\color{blue}{\spadesuit}$ &
          44.16/\textbf{21.28}/40.90 $\color{blue}{\spadesuit}$ \ \ &
          \textbf{45.09}/\textbf{19.56}/27.42 $\color{blue}{\blacklozenge}$ \ \  & 
          42.86/\textbf{19.71}/34.80 $\color{blue}{\blacklozenge}$ \ \ \ & 
          49.30/25.60/47.70 $\color{blue}{\star}$  \\
          FT (repr.). & 
          \textbf{45.47}/\textbf{22.40}/\textbf{37.42} & 
          \textbf{44.30}/21.14/\textbf{41.22} & 
          44.96/19.00/\textbf{27.74}& 
          \textbf{43.26}/19.38/\textbf{34.89} &
          \textbf{53.02}/\textbf{28.30}/\textbf{48.73}   
          \\
          \hline
         \multicolumn{6}{l}{{\textbf{PrefixTune (PT)}}} \\
         \hline

        PT \textbf{[*]}    & 
         43.80/\textbf{\underline{20.93}}/36.05 $\color{blue}{\clubsuit}$ &
         \multicolumn{1}{c}{-} & 
         \multicolumn{1}{c}{-} &  
         \multicolumn{1}{c}{-} & \multicolumn{1}{c}{-} \\
         
        PT (repr.)  & 
        43.43/20.37/35.47 & 
        42.50/19.52/39.08 &  
        42.38/16.31/24.57 & 
        39.26/15.58/28.27 &  
        52.12/26.52/48.05
        \\
        UniBlock & 
         43.49/20.58/35.80 &
         42.43/19.38/39.15 &
         42.81/16.83/24.87 & 
         39.34/15.55/28.75&
         52.42/27.70/48.12
         \\
        HierBlock & \textbf{\underline{43.91}}/20.83/\textbf{\underline{36.38}} &
         \textbf{\underline{43.33}}/\textbf{\underline{20.27}}/\textbf{\underline{40.12}} &
         \textbf{\underline{43.16}}/\textbf{\underline{16.96}}/\textbf{25.73} & 
         \textbf{\underline{40.03}}/\textbf{\underline{15.90}}/\textbf{\underline{30.15}}& 
         \textbf{\underline{52.68}}/\textbf{\underline{27.88}}/\textbf{\underline{48.56}} 
         \\
  \hline
    \end{tabular}
    \vspace{-0.3cm}
    \caption{\small \textbf{Summarization}: \textbf{Prefix blocking} experiment results comparing against finetuning and prefix-tuning. Top block are best reported scores from corresponding papers \textbf{[*]}: $\color{blue}{\spadesuit}$ \citep{lewis-etal-2020-bart}, $\color{blue}{\clubsuit}$ \citep{prefixtune}, $\color{blue}{\star}$\citep{samsumbartlarge}, $\color{blue}{\blacklozenge}$ \citep{pegasus}. (repr.) denotes our replication of finetuned and prefixtuned BART models. Bottom block are our prefixtuned models: Uniform (UniBlock) and Hierarchical (HierBlock) prefix blocking represent models which use prefix-blocking at different layers ($\S$~\ref{prefixblocking}). All models use BART-large. The best finetune (top block) models are \textbf{bolded}, best prefixtune models (bottom block) are further \underline{\textbf{underlined}}. All hyper-parameters are summarized in Appendix Table~\ref{apptable:hyperparameters}.}
    \label{tab:res1}
\end{table*}

\noindent{\textbf{Datasets.}}
\label{datasetsintext}
We conduct experiments across six datasets on two tasks: abstractive summarization and data-to-text (S2T) generation. We present a summary of the datasets in Table~\ref{tab:datasets} and provide more details about the datasets in Appendix~\ref{app:dataset_details}.

\noindent{\textbf{Metrics.}} For all the tasks and datasets we use the \textit{n}-gram match metrics: ROUGE-1/2/L \citep{lin-2004-rouge} for summarization. We use BLEU \citep{bleu}, NIST \citep{nist}, METEOR \citep{meteor}, ROUGE-L, TER, Movers \citep{moverscore} and BERTScore \citep{bertscore} for data-to-text tasks. We also report human evaluations analysis.

\section{Experiment Results}
\label{experimentresults}
\subsection{Are all prefix-parameters useful?}
\label{exp:prefixanalysis}
\noindent\textbf{Finding:} 
Prefix-tuning models encode diverse but task specific features at each layer differently, while the top-layer prefixes encode abstract features.

\noindent\textbf{Analysis:} 
Earlier work \citep{jiang-etal-2021-enriching,Elazar2021AmnesicPB,beyondbert} suggests that 
some layers of the transformers are better than others at producing representations that are useful for a given task.
To investigate if similar patterns show up in prefix-tuned models, we take XSum dataset and train models with prefix parameters only at the top layers, the bottom layers, and at a single layer.
\begin{table}[h]%
\centering
\parbox{0.2\textwidth}{
\begin{footnotesize}
    \begin{tabular}{@{}l@{}c@{} }
        \hline 
        \footnotesize Layers \ & Rouge-1/2/L \\
         \hline
         \hline
        Top (8-12) \ \ &  42.3/18.1/33.4  \\
        Low (1-7) &  35.7/13.1/26.9   \\
        \hline
        All (1-12) &  42.6/19.3/34.2   \\
        \hline
    \end{tabular}

\end{footnotesize}
\vspace{-0.3cm}
\caption{\footnotesize Validation Rouge scores of prefix-tuned models on XSum using only top/low layers.}
\label{tab:minirouge}
}
\qquad
\begin{minipage}[c]{0.2\textwidth}%
\centering
    \begin{tikzpicture}[yscale=0.32,xscale=0.40]
\begin{axis}[
font=\fontsize{15}{0}\selectfont,
xlabel=Layers,
]
\addplot[color=red,mark=x] coordinates {
(1,	27.8497)
(2,27.9531)
(3,28.5052)
(4,28.5875)
(5,28.5239)
(6,28.7596)
(7,29.0929)
(8,29.38)
(9,33.0148)
(10,33.1243)
(11,33.4351)
(12,33.8272)
};
\end{axis}
\end{tikzpicture}
\vspace{-0.4cm}

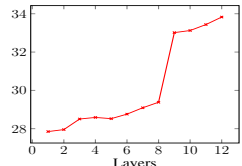
\captionof{figure}{\footnotesize Validation \textbf{Rouge-L} on single-layer prefix-tuning with XSum.}
\label{fig:singlelayers}
\end{minipage}
\end{table}


We show layer-specific prefix-tuned models' validation performance results in Table~\ref{tab:minirouge}. The 'Top' layers model is tuned with only the top-layer prefix parameters (i.e., top 4 layers have additional prefix parameters), the 'Low' layers model uses only the lower-layer prefix-parameters (i.e., bottom 7 layers have additional prefix parameters) and 'All' layers prefix parameters is same as baseline prefix-tuning. On inspection, we see a  moderate/huge performance gap between the models trained with top/lower layers, while we obtain the best performance when we tune all-layer prefix parameters.  We see similar patterns on the SAMSum dialog summarization and E2E structure to text generation tasks (in Appendix~\ref{app:layerwiseprefix}). We also build models when prefix parameters are used at a single layer of the network.  On single layers in Figure~\ref{fig:singlelayers}, all layers contribute to the performance of the summarization task, the top layer prefixes perform the best suggesting they might be encoding summary related abstract information.

\subsection{Are hierarchical prompts effective?} 
\label{blockingresults}
\noindent\textbf{Finding:} Hierarchical design of prefix parameters can yield more robust information transfer in text generation outperforming baseline prefix-tuning.


\noindent\textbf{Analysis:} 
To simulate learning the discourse related representations we bias prefix parameters with a structure of  input documents (as discussed in $\S$\ref{discourse}) and experiment with two hierarchical structures: \textit{uniform} (UniBlock) and \textit{hierarchical} (HierBlock) from $\S$~\ref{prefixblocking}. 
In Table~\ref{tab:res1} we report the performance of our models in comparison to fine-tuning and baseline prefix-tuning on abstractive summarization tasks.  Our results indicate that prefix-blocking
models improve over the baseline prefix-tuning 
on all summarization tasks by up to +1.1 ROUGE-L score overall. Especially for Wikihow, which are considered long document summarization task, we observe up to +2 ROUGE-L score improvement. 
We further observe that hierarchical blocking on prefixes also helps for data-to-text tasks, though the performance impact of structural bias is more prominent in summarization tasks. We show detailed results of data-to-text tasks and provide samples of generated outputs in Appendix~\ref{app:hierpromptdesign}. 


\begin{figure}[t]
    \centering
    {\includegraphics[scale=0.5,height=4cm]{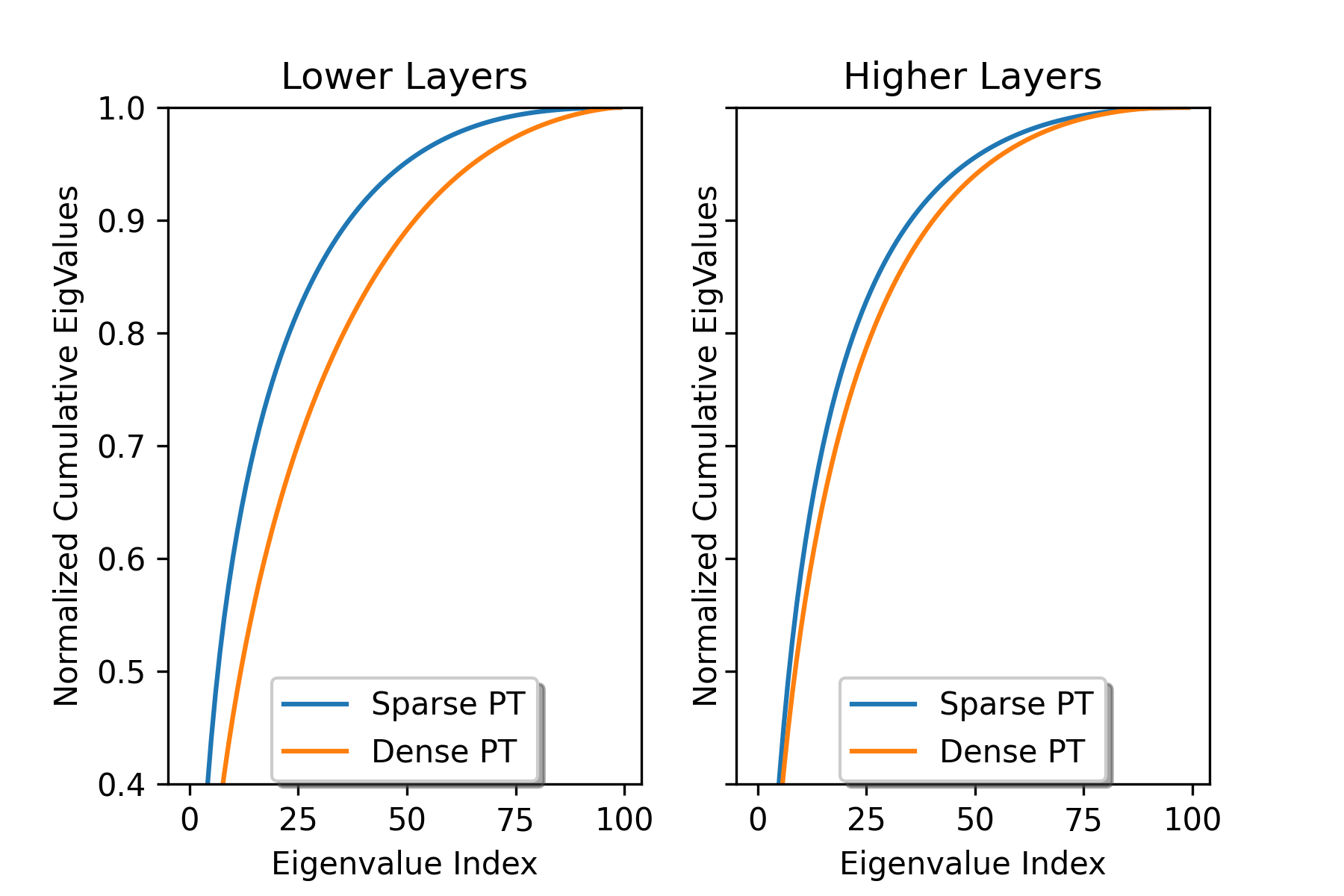} }
    \vspace{-0.4cm}
    \caption{\small Spectrum analysis of the self-attention matrix comparing the baseline (Dense) and our Sparse Prefix-Tuned (PT) transformer
model zooming in on prefix parameters of size 100. The Y-axis is the normalized cumulative singular value of the self-attantion matrix \textbf{A}, and the X-axis the index of largest eigenvalue. The results are based on BART-Large on XSum dataset. The left plots the averages of all \textbf{A} on the lower layers, while right plots averages over higher layers. 
}%
    \label{fig:eigen}%
\end{figure}

\subsection{Does sparse attention help prefix-tuning?}
\label{doessparseattentionlhelp}
\noindent\textbf{Finding:} With hierarchically structured sparsity training, prefix tuning show more sparse patterns at the lower layers. Sparse prefix parameters at lower layers, and dense at higher layers enable more efficient tuning of the prefix-parameters.

\noindent\textbf{Spectrum Analysis (Statistical Proof):} 
To investigate if our sparse models do in fact learn sparse representations, we conduct spectrum analysis on the encoder attention matrix \textbf{A} zooming in on the prefix parameters\footnote{A similar spectrum analysis has been used to prove the sparsity of the attention matrix in Linformer \citep{linformer}, a sparse transformer.}. To analyze the variation of attention scores we calculate the principal components of the attention scores of prefix parameters\footnote{Eigenvalues capture the variation of the
attention scores distribution along different principal components.} and plot in Figure~\ref{fig:eigen}. We observe that the spectrum distribution of prefixes in lower layers is more skewed than in
higher layers, meaning that, in lower layers, more information is concentrated in the largest singular
values and the rank of \textbf{A} is lower. In summary, with sparse attention at the lower layers and dense attention at the top layers, the prefix-tuned models enables encoding important features necessary for the factual and consistent summarization. Details on spectrum analysis are provided in Appendix~\ref{app:spectrum}. Next, we empirically support this statistical proof.

\begin{table}[t!]
    \centering
    \fontsize{8}{9}\selectfont
        \begin{tabular}{@{}l@{}c@{}c@{}c@{}c@{}c@{}}
        \hline 
        \textbf{Method} & \textbf{XSum} \ \ & \textbf{CNN} \ \ & \textbf{PubMed} \ \ & \textbf{Wikihow} \ \ & \textbf{SAMSum}\\
        \hline
        \multicolumn{6}{l}{\textit{\textbf{Finetune}}} \\
        \hline
        Dense   & 
        \textbf{37.42} & 
        41.22 & 
        27.42 & 
        34.89 & 
        48.73 \\
        (\textit{a}) TruncSA  & 
        37.02& 
        39.96& 
        26.37& 
        35.58& 
        48.12\\
        (\textit{b}) SoftSA  & 
        37.23& 
        39.67& 
        26.26& 
        32.53& 
        48.45\\
        \hline
        \multicolumn{6}{l}{\textit{\textbf{Prefix-tune}}} \\
        \hline
        Dense   & 
        35.47 & 
        39.08& 
        24.57& 
        28.27& 
        48.05\\
        (\textit{a}) TruncSA   & 
        35.39 & 
        39.52& 
        25.61& 
        28.94& 
        47.88\\
        (\textit{b}) SoftSA   & 
        35.94 & 
        39.24& 
        & 
        28.94& 
        47.35\\
        (\textit{c}) HTruncSA \ \ \    &  
        36.42&  
        40.00&  
        25.28& 
        30.02& 
        48.00\\
        (\textit{d}) HSoftSA    &  
        36.13&  
        39.83& 
        24.90& 
        30.01& 
        48.33\\
    \hline
    \end{tabular}
    \vspace{-0.2cm}
    \caption{\small Sparse Attention experiment \textbf{ROUGE-L} results on Finetuning, and Prefix-tuning using dense and soft sparse attention designs in $\S$\ref{doessparseattentionlhelp}. 
    The best results accross finetune models are \textbf{bolded}, the best prefixtune models are also \underline{\textbf{underlined}}. Full results are included in Appendix~\ref{apptab:sparse}.}
    \label{tab:sparse}
\end{table}

\noindent\textbf{Sparsity Analysis:} 
We investigate the impact of sparsity on the performance of the prefix-tuning models. For a fair comparison, we also apply attention sparsity on the finetuned models. We build prefix-tuning models with (\textit{a}) Truncated Sparse Attention (TruncSA), (\textit{b}) Soft Sparse Attention (SoftSA), (\textit{c}) Hierarchical TruncSA (HTruncSA), with top-\textit{p} sparsity at the lower layers, and dense attention at the top layers, (\textit{d}) Hierarchical Soft Sparse Attention (HSoftSA), with soft sparse attention at the lower layers but dense at top layers. 

We show the ROUGE-L results in Table~\ref{tab:sparse}. We observe that when sparsity is used on the prefix-parameters, the prefix-tuned models 
outperform baseline all-dense prefix-tuning models on all datasets. The performance improvements are more pronounced on long document summarization tasks such as Wikihow, reaching close to 2.0 ROUGE-L improvements. 
Comparing all layers sparse models of (\textit{a}) and (\textit{b}) to hierarchically biased sparsity models of (\textit{c}) and (\textit{d}), we observe improvements with the hierarchically structured sparse prefix-tuning models. More details on quantitative analysis are provided in Appendix~\ref{app:spectrum} and Table~\ref{apptab:sparse}.

\begin{table}[t]
    \centering
    \fontsize{8}{9}\selectfont
    \begin{tabular}{@{}lcc@{}}
        \hline 
        Dataset & \textbf{HierBlock} & \textbf{HierBlock+SoftSA} \\
        \hline
        \textbf{Summarization} & R1/R2/R-L & R1/R2/R-L \\
        \hline
        XSum &  43.91/20.83/36.38 &  \textbf{44.00}/\textbf{20.93}/\textbf{36.59}\\
        CNN/DM & \textbf{43.33}/20.27/\textbf{40.12} & \textbf{43.33}/\textbf{20.31}/40.10\\
        PubMED & \textbf{43.16}/\textbf{16.83}/24.87 & 42.96/16.64/\textbf{25.00}\\
        Wikihow & \textbf{39.34}/\textbf{15.55}/\textbf{28.75} &  39.30/15.42/28.67  \\
        SAMSum & 52.58/27.58/48.42 & \textbf{52.83}/\textbf{27.94}/\textbf{48.72}\\
        \hline
        \textbf{Data-to-Text} & \\
        \hline
        & BLEU/R-L/CiDER & BLEU/R-L/CiDER\\
        E2E & 67.2/69.1/2.35 & \textbf{68.0}/\textbf{69.6}/\textbf{3.38}\\
        \hline
        & BLEU/MET/TER↓ & BLEU/MET/TER↓\\
        DART & \textbf{46.6}/0.39/\textbf{0.45}& 46.0/0.39/0.46 \\
        \hline
    \end{tabular}
    \vspace{-0.3cm}
    \caption{\small What happens when we introduce sparsity to hierarchically blocked prompt design? Results comparing dense and sparse prefix-tuning with structurally biased prefix design (via hierarchical blocking) on various text generation tasks. The best results across two models are \textbf{bolded}.}
    \label{tab:res3}
    \vspace{-0.3cm}
\end{table}

\subsection{Does sparsity on hierarchically blocked prefixes further improve performance?}
\label{sparsityinhierblock}
\noindent{\textbf{Finding:}} 
The most performance gains are obtained when
sparsity constraints are applied on the hierarchically blocked prefixes (Table~\ref{tab:res3}).

\noindent{\textbf{Analysis:}}
Recall from the earlier discussions in $\S$\ref{blockingresults} that, if applying blocking on the lower layered prefixes, while letting all tokens attend to all prefixes at the top layers (HierBlock models) can improve performance. On separate set of ablations in $\S$\ref{doessparseattentionlhelp}, we also observe that if we introduce sparsity at different layers of the network, the sparse parameters influence the performance compared to the dense prefix tuned parameters at all layers. We now introduce sparsity on the hierarchically blocked prefix-models, combining the best hierarchically blocked prefix-tuned models with the sparse attention. 

In Table~\ref{tab:res3} we show results of our hierarchical prefix blocking (HierBlock) model against hierarchical prefix blocking model with soft sparse attention (HierBlock+SoftSA). To build the HierBlock+SoftSA models, we apply soft sparsity at the lower layers with blocked prefix parameters, while the top layers use dense prefixes with all tokens attending to all prefixes. In Table~\ref{tab:res3} we repeat the results of the last row from Table~\ref{tab:res1} for easy comparison.  We observe performance improvements on summarization tasks where the output summaries are shorter, (e.g., XSum SAMSum) and less on the longer summaries (e.g., Pubmed, Wikihow). On the data-to-text generation tasks the sparsity on hierarchical blocking only improves on E2E, though both HierBlock and HierBlock+SoftSA perform better than baseline prefix-tuning models (see App. Table~\ref{apptab:table2textsparsity}). More details are provided in Appendix~\ref{app:sparsityinhierblock}. Our analysis suggests that discourse aware design can improve prefix-tuning models when the output generations are short (<100 tokens). 


\begin{table}
\setlength{\tabcolsep}{3pt}
    \centering
    \small
        \begin{tabular}{rr|cccc}
        &  & \multicolumn{4}{c}{Wins \% matches} \\
        \multicolumn{2}{r|}{\textbf{Faithfullness}} & {PT} & {HSoftSA} & {HB+SoftSA} & {HB} \\
        \midrule
        \parbox[t]{2mm}{\multirow{4}{*}{\rotatebox[origin=c]{90}{Loses \%}}}
        
        & PT & & \win{50.0} & \win{{\textbf{64.3}}}& \win{{50.0}} \\[-0.25mm] 
        
        & HSoftSA & \lose{50.0} &  & \win{{\textbf{60.0}}} & \win{{\textbf{60.0}}}   \\[-0.25mm] 
        
        & HB+SoftSA & \lose{35.7} & \lose{40.0} & & \win{53.9} \\[-0.25mm] 
        & HB & \lose{50.0} & \lose{40.0} & \lose{46.1} & \\[-0.25mm] 
        \end{tabular}
        \begin{tabular}{rr|cccc}
        &  & \multicolumn{4}{c}{Wins \% matches} \\
        \multicolumn{2}{r|}{\textbf{Overall}} & {PT} & {HSoftSA} & {HB+SoftSA} & {HB} \\
        \midrule
        \parbox[t]{2mm}{\multirow{4}{*}{\rotatebox[origin=c]{90}{Loses \%}}}
        & PT & & \win{\textbf{67.5}} & \win{{\textbf{62.0}}}& \win{48.1} \\[-0.25mm] 
        & HSoftSA & \lose{32.5} &  & \win{{\textbf{63.2}}} & \win{\textbf{57.9}}   \\[-0.25mm] 
        & HB+SoftSA & \lose{38.0} & \lose{36.8} & & \win{{50.0}} \\[-0.25mm] 
        & HB & \lose{51.9} & \lose{42.1} & \lose{50.0} & \\[-0.25mm] 
        \end{tabular}
    \vspace{-0.1cm}
    \caption{Human evaluation results on {\em Faithfullness} (top) and {\em Overall} (bottom) ratings. PT: Prefixtune, HSoftSA: Hierarchical Soft Attention, HB: HierBlock, HB+SoftSA: HierBlock with Soft Sparse Attention. Bold win \%s indicate significance ($p<.05$).}
    \label{tab:human_eval_pt}
\end{table}

\subsection{Do human evals. support our claims?}
\label{humanevals}
\noindent{\textbf{Finding:}} Humans generally prefer generated text from hierarchically blocked prefix-tuned models over all other models, find overall quality of generations indistinguishable from fine-tuning.

\noindent{\textbf{Analysis:}} To evaluate the generated text from our proposed methods against baseline models, we ask human annotators to rate generations on five criteria: \textit{faithfulness} (consistent with the context), \textit{relevance} (captures key-points), \textit{grammaticality}, \textit{coherence} (form a cohesive whole), and \textit{overall quality} (helpful, informative). Table~\ref{tab:human_eval_pt} shows the results of the study on faithfulness, and overall metrics. 
The columns show the percentage of wins of the model
against its opponent on a given row. Our Hierarhical Blocking (HierBlock) and  Hierarchical Soft Sparse Attention (HSoftSA) 
models beat prefix-tuning and HierBlock significantly ($p < .05$) beats most of our sparse models on all axes including factuality. Especially, on relevance metrics, all our models perform better than prefix-tuning and even outperforming fintuning.  More details about the evaluation setup as well as results on all the criteria comparing against fine-tuning and prefix-tuning can be found in Appendix~\ref{app:humanevals}. In Table~\ref{apptab:humandetails} we provide comparisons with fine-tuning 
and observe that HierBlock models perform as good as finetuning on all criteria.

\subsection{Which structural features are harder to transfer in low-resource settings?}
\label{lowdatasetting}
\noindent{\textbf{Finding:}} In low-resource settings, hierarchically designed sparse prefix parameters can efficiently transfer knowledge and represent the semantics and structure of input text yielding more accurate output generations. 

\noindent{\textbf{Analysis:}} We simulate a low-resource setting by randomly sampling \textit{k}\% (k={5,10,25,50}) from the training dataset of two summarization tasks: XSum on news, and Wikihow on DIY domains (see train data sizes in Table~\ref{tab:datasets}). 
We use the same hyperparameter settings as our previous models detailed in $\S$~\ref{setup}. We compare our approach to finetuning and prefix-tuning under low-resource settings.


\begin{wrapfigure}{r}{0.2\textwidth}
  \vspace{-12pt}
     \begin{tikzpicture}[yscale=0.40, xscale=0.40] 
     \begin{axis}[
     font=\fontsize{15}{0}\selectfont,
     xmax=50,xmin=0,
     ymin= 27.5,ymax=31.5,
     xlabel=\emph{\% train samples}
     xtick={5,10,25,50},
     ytick={27,28,29,30,31},
     legend columns=2, 
     legend style={at={(0.5,1.15)},anchor=north},line width=1pt,
     ]
     \addplot [green!60!black,mark=square] coordinates{ (5,28.45) (10,29.09) (25,30.53) (50,31.05)};
     \addplot [olive,mark=x] coordinates{ (5,27.87) (10,28.79) (25,29.78) (50,30.27)};
     \addplot [blue,mark=*] coordinates{ (5,28.54) (10,29.21) (25,30.09) (50,30.68)};
     \addplot [red,mark=o] coordinates{ (5,28.94) (10,29.71) (25,30.38) (50,30.60)};
          \legend{\emph{Finetune},\emph{Prefix-tune},\emph{HierBlock},\emph{HierBlock+SA}}
     \end{axis} 
     \end{tikzpicture}
\caption{\small Average ROUGE-L scores on \textbf{low-resource settings}.} 
\label{fig:low_small}
  \vspace{-10pt}
\end{wrapfigure}
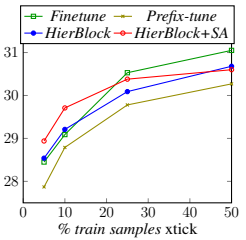

In Figure~\ref{fig:low_small} on the right, we plot ROUGE-L averaging scores of models trained on XSUM and Wikihow. Our structured prefix-tuned models, HierBlock (\textcolor{blue}{blue}) and its sparse extension which uses sparse features, HierBlock+SA (\textcolor{red}{red})
outperforms fine-tuned (\textcolor{green}{green}) and prefix-tuned models (\textcolor{olive}{olive}), while using the same number of parameters in low resources settings (when <50\% training samples are used). 
Although HierBlock models show consistent performance, on low-resource settings HierBlock-SA performance is more stable.
(See Appendix~\ref{app:lowdatasetting} for more details.)

\section{Conclusion and Limitations}
We have described simple but effective prompt design options for prefix-tuning of text generation tasks. We enrich prefix parameters with structural biases by way of: prefix-blocking at different layers of the network, sparsity on prefix-parameters and an ensemble of both biases. We show with quantitative and human evaluations on metrics such as coherence and faithfullness that discourse aware prefix designs outperforms baseline prefix-tuning across all text generation tasks even at low data settings.

We note a few limitations of our work: (1) our experiments are limited by available datasets, and only evaluated on limited closed domain text generation tasks; (2) we focused on efficient prefix-tuning, while ensemble of different efficient tuning models can boost performance even further; (3) we conduct experiments with ${\small \sim}$300M parameter models as in past work, but it will be valuable for future work to scale to larger models which may exhibit more coherent generations.


\section{Ethics Statement}
\label{ethics}
In this work we apply several changes to the state-of-the-art encoder-decoder modeling architecture and build several models to benchmark our new architecture with baseline architectures on several open source text generation datasets. 

\paragraph{Intended use.}
Our architecture is designed to build models of abstractive document summarization and table summarization. Potentially our architecture could be used to train models for summarizing any type of datasets (e.g., any documents, textual conversational dialogues, blog posts, reports, meetings, legal forms, etc.) to further improve the productivity and efficiency of the users in their daily activities without needing to read/listen to long documents/conversations/meetings.

\paragraph{Failure mode.}
Even though our models yield factually consistent summaries, as judged by us and raters, they can still generate factually inconsistent summaries or sometimes hallucinate information that the source document does not include. This might be due to the bias or noise in the training data. Model builders wanting to use our architecture to build models on their datasets should build models with consideration of intellectual properties and privacy rights.

\paragraph{Misuse Potential.} We note the models to be built with our architecture should be used with careful consideration especially if used to build summarization models. The generated summaries produced by our models are not controlled and use generative approaches, therefore, they could generate unreliable text. Researchers working on abstractive summarization should focus on generating factually correct, ethical and reliable text. If our models are trained on news datasets, a careful consideration should be made on factuality of the generated text and measures have been taken to prevent model hallucinations.

\bibliography{custom}

\begin{thebibliography}{65}
\expandafter\ifx\csname natexlab\endcsname\relax\def\natexlab#1{#1}\fi

\bibitem[{Barzilay and Lapata(2008)}]{coherence2}
Regina Barzilay and Mirella Lapata. 2008.
\newblock \href {https://aclanthology.org/J08-1001.pdf} {Modeling local
  coherence: An entity-based approach}.
\newblock \emph{Computational Linguistics}, 34(1).

\bibitem[{Barzilay and Lee(2004)}]{Barzilay2004CatchingTD}
Regina Barzilay and Lillian Lee. 2004.
\newblock \href {https://aclanthology.org/N04-1015.pdf} {Catching the drift:
  Probabilistic content models, with applications to generation and
  summarization}.
\newblock In \emph{Proceedings of the 2004 Conference of the North American
  Chapter of the Association for Computational Linguistics: Human Language
  Technologies}.

\bibitem[{Belz and Reiter(2006)}]{nist}
Anja Belz and Ehud Reiter. 2006.
\newblock \href {https://aclanthology.org/E06-1040} {Comparing automatic and
  human evaluation of {NLG} systems}.
\newblock In \emph{11th Conference of the {E}uropean Chapter of the Association
  for Computational Linguistics}, pages 313--320, Trento, Italy. Association
  for Computational Linguistics.

\bibitem[{Brown et~al.(2020)Brown, Mann, Ryder, Subbiah, Kaplan, Dhariwal,
  Neelakantan, Shyam, Sastry, Askell, Agarwal, Herbert-Voss, Krueger, Henighan,
  Child, Ramesh, Ziegler, Wu, Winter, Hesse, Chen, Sigler, Litwin, Gray, Chess,
  Clark, Berner, McCandlish, Radford, Sutskever, and Amodei}]{gpt3}
Tom Brown, Benjamin Mann, Nick Ryder, Melanie Subbiah, Jared~D Kaplan, Prafulla
  Dhariwal, Arvind Neelakantan, Pranav Shyam, Girish Sastry, Amanda Askell,
  Sandhini Agarwal, Ariel Herbert-Voss, Gretchen Krueger, Tom Henighan, Rewon
  Child, Aditya Ramesh, Daniel Ziegler, Jeffrey Wu, Clemens Winter, Chris
  Hesse, Mark Chen, Eric Sigler, Mateusz Litwin, Scott Gray, Benjamin Chess,
  Jack Clark, Christopher Berner, Sam McCandlish, Alec Radford, Ilya Sutskever,
  and Dario Amodei. 2020.
\newblock \href
  {https://proceedings.neurips.cc/paper/2020/file/1457c0d6bfcb4967418bfb8ac142f64a-Paper.pdf}
  {Language models are few-shot learners}.
\newblock In \emph{Advances in Neural Information Processing Systems},
  volume~33, pages 1877--1901. Curran Associates, Inc.

\bibitem[{Chen and Yang(2020)}]{samsumbartlarge}
Jiaao Chen and Diyi Yang. 2020.
\newblock \href {https://doi.org/10.18653/v1/2020.emnlp-main.336} {Multi-view
  sequence-to-sequence models with conversational structure for abstractive
  dialogue summarization}.
\newblock In \emph{Proceedings of the 2020 Conference on Empirical Methods in
  Natural Language Processing (EMNLP)}, pages 4106--4118, Online. Association
  for Computational Linguistics.

\bibitem[{Cohan et~al.(2018)Cohan, Dernoncourt, Kim, Bui, Kim, Chang, and
  Goharian}]{cohan-etal-2018-discourse}
Arman Cohan, Franck Dernoncourt, Doo~Soon Kim, Trung Bui, Seokhwan Kim, Walter
  Chang, and Nazli Goharian. 2018.
\newblock \href {https://doi.org/10.18653/v1/N18-2097} {A discourse-aware
  attention model for abstractive summarization of long documents}.
\newblock In \emph{Proceedings of the 2018 Conference of the North {A}merican
  Chapter of the Association for Computational Linguistics: Human Language
  Technologies, Volume 2 (Short Papers)}, pages 615--621, New Orleans,
  Louisiana. Association for Computational Linguistics.

\bibitem[{Cui et~al.(2019)Cui, Li, Chen, and Zhang}]{cui-etal-2019-fine}
Baiyun Cui, Yingming Li, Ming Chen, and Zhongfei Zhang. 2019.
\newblock \href {https://doi.org/10.18653/v1/D19-1361} {Fine-tune {BERT} with
  sparse self-attention mechanism}.
\newblock In \emph{Proceedings of the 2019 Conference on Empirical Methods in
  Natural Language Processing and the 9th International Joint Conference on
  Natural Language Processing (EMNLP-IJCNLP)}, pages 3548--3553, Hong Kong,
  China. Association for Computational Linguistics.

\bibitem[{Cui et~al.(2021)Cui, Wu, Liu, Yang, and
  Zhang}]{cui-etal-2021-template}
Leyang Cui, Yu~Wu, Jian Liu, Sen Yang, and Yue Zhang. 2021.
\newblock \href {https://doi.org/10.18653/v1/2021.findings-acl.161}
  {Template-based named entity recognition using {BART}}.
\newblock In \emph{Findings of the Association for Computational Linguistics:
  ACL-IJCNLP 2021}, pages 1835--1845, Online. Association for Computational
  Linguistics.

\bibitem[{Dai et~al.(2021)Dai, Cahyawijaya, Liu, and
  Fung}]{dai-etal-2021-multimodal}
Wenliang Dai, Samuel Cahyawijaya, Zihan Liu, and Pascale Fung. 2021.
\newblock \href {https://doi.org/10.18653/v1/2021.naacl-main.417} {Multimodal
  end-to-end sparse model for emotion recognition}.
\newblock In \emph{Proceedings of the 2021 Conference of the North American
  Chapter of the Association for Computational Linguistics: Human Language
  Technologies}, pages 5305--5316, Online. Association for Computational
  Linguistics.

\bibitem[{Devlin et~al.(2019)Devlin, Chang, Lee, and Toutanova}]{bert}
Jacob Devlin, Ming-Wei Chang, Kenton Lee, and Kristina Toutanova. 2019.
\newblock \href {https://aclanthology.org/N19-1423.pdf} {Bert: Pre-training of
  deep bidirectional transformers for language understanding}.
\newblock In \emph{Proceedings of the 2019 Conference of the North American
  Chapter of the Association for Computational Linguistics: Human Language
  Technologies}.

\bibitem[{Dong et~al.(2021)Dong, Mircea, and Cheung}]{dong-etal-2021-discourse}
Yue Dong, Andrei Mircea, and Jackie Chi~Kit Cheung. 2021.
\newblock \href {https://doi.org/10.18653/v1/2021.eacl-main.93}
  {Discourse-aware unsupervised summarization for long scientific documents}.
\newblock In \emph{Proceedings of the 16th Conference of the European Chapter
  of the Association for Computational Linguistics: Main Volume}, pages
  1089--1102, Online. Association for Computational Linguistics.

\bibitem[{Du{\v{s} }ek et~al.(2019)Du{\v{s} }ek, Howcroft, and Rieser}]{e2e}
Ond{\v{r} }ej Du{\v{s} }ek, David~M. Howcroft, and Verena Rieser. 2019.
\newblock \href {https://doi.org/10.18653/v1/W19-8652} {Semantic noise matters
  for neural natural language generation}.
\newblock In \emph{Proceedings of the 12th International Conference on Natural
  Language Generation}, pages 421--426, Tokyo, Japan. Association for
  Computational Linguistics.

\bibitem[{Elazar et~al.(2021)Elazar, Ravfogel, Jacovi, and
  Goldberg}]{Elazar2021AmnesicPB}
Yanai Elazar, Shauli Ravfogel, Alon Jacovi, and Yoav Goldberg. 2021.
\newblock \href
  {https://direct.mit.edu/tacl/article/doi/10.1162/tacl_a_00359/98091/Amnesic-Probing-Behavioral-Explanation-with}
  {Amnesic probing: Behavioral explanation with amnesic counterfactuals}.
\newblock \emph{Transactions of the Association for Computational Linguistics},
  9:160--175.

\bibitem[{Fabbri et~al.(2019)Fabbri, Li, She, Li, and
  Radev}]{fabbri-etal-2019-multi}
Alexander Fabbri, Irene Li, Tianwei She, Suyi Li, and Dragomir Radev. 2019.
\newblock \href {https://doi.org/10.18653/v1/P19-1102} {Multi-news: A
  large-scale multi-document summarization dataset and abstractive hierarchical
  model}.
\newblock In \emph{Proceedings of the 57th Annual Meeting of the Association
  for Computational Linguistics}, pages 1074--1084, Florence, Italy.
  Association for Computational Linguistics.

\bibitem[{Fan et~al.(2018)Fan, Lewis, and Dauphin}]{Fan2018HierarchicalNS}
Angela Fan, Mike Lewis, and Yann Dauphin. 2018.
\newblock \href {https://doi.org/10.18653/v1/P18-1082} {Hierarchical neural
  story generation}.
\newblock In \emph{Proceedings of the 56th Annual Meeting of the Association
  for Computational Linguistics (Volume 1: Long Papers)}, pages 889--898,
  Melbourne, Australia. Association for Computational Linguistics.

\bibitem[{Gliwa et~al.(2019)Gliwa, Mochol, Biesek, and Wawer}]{gliwa2019samsum}
Bogdan Gliwa, Iwona Mochol, Maciej Biesek, and Aleksander Wawer. 2019.
\newblock \href {https://doi.org/10.18653/v1/D19-5409} {{SAMS}um corpus: A
  human-annotated dialogue dataset for abstractive summarization}.
\newblock In \emph{Proceedings of the 2nd Workshop on New Frontiers in
  Summarization}, pages 70--79, Hong Kong, China. Association for Computational
  Linguistics.

\bibitem[{Gu et~al.(2021)Gu, Yoo, and Lee}]{promptdesign}
Xiaodong Gu, Kang~Min Yoo, and Sang{-}Woo Lee. 2021.
\newblock \href {http://arxiv.org/abs/2111.02643} {Response generation with
  context-aware prompt learning}.
\newblock \emph{ArXiv}.

\bibitem[{Han et~al.(2021)Han, Zhao, Ding, Liu, and Sun}]{hanpromptcomp}
Xu~Han, Weilin Zhao, Ning Ding, Zhiyuan Liu, and Maosong Sun. 2021.
\newblock \href {http://arxiv.org/abs/2105.11259} {{PTR:} prompt tuning with
  rules for text classification}.
\newblock \emph{ArXiv}, abs/2105.11259.

\bibitem[{Hermann et~al.(2015)Hermann, Kocisky, Grefenstette, Espeholt, Kay,
  Suleyman, and Blunsom}]{hermann2015teaching}
Karl~Moritz Hermann, Tomas Kocisky, Edward Grefenstette, Lasse Espeholt, Will
  Kay, Mustafa Suleyman, and Phil Blunsom. 2015.
\newblock \href
  {https://proceedings.neurips.cc/paper/2015/file/afdec7005cc9f14302cd0474fd0f3c96-Paper.pdf}
  {Teaching machines to read and comprehend}.
\newblock In \emph{Advances in Neural Information Processing Systems},
  volume~28. Curran Associates, Inc.

\bibitem[{Holtzman et~al.(2020)Holtzman, Buys, Du, Forbes, and Choi}]{nucleus}
Ari Holtzman, Jan Buys, Li~Du, Maxwell Forbes, and Yejin Choi. 2020.
\newblock \href {https://openreview.net/forum?id=rygGQyrFvH} {The curious case
  of neural text degeneration}.
\newblock In \emph{International Conference on Learning Representations}.

\bibitem[{Houlsby et~al.(2019)Houlsby, Giurgiu, Jastrzebski, Morrone,
  de~Laroussilhe, Gesmundo, Attariyan, and Gelly}]{adapters}
Neil Houlsby, Andrei Giurgiu, Stanislaw Jastrzebski, Bruna Morrone, Quentin
  de~Laroussilhe, Andrea Gesmundo, Mona Attariyan, and Sylvain Gelly. 2019.
\newblock \href {http://arxiv.org/abs/1902.00751} {Parameter-efficient transfer
  learning for {NLP}}.
\newblock In \emph{International Conference of Machine Learning}.

\bibitem[{Jang et~al.(2016)Jang, Gu, and Poole}]{jang2017categorical}
Eric Jang, Shixiang Gu, and Ben Poole. 2016.
\newblock \href {http://arxiv.org/abs/1611.01144} {Categorical
  reparameterization with gumbel-softmax}.
\newblock In \emph{International Conference on Learning Representations}.

\bibitem[{Jawahar et~al.(2019)Jawahar, Sagot, and
  Seddah}]{jawahar-etal-2019-bert}
Ganesh Jawahar, Beno{\^\i}t Sagot, and Djam{\'e} Seddah. 2019.
\newblock \href {https://aclanthology.org/P19-1356.pdf} {What does {BERT} learn
  about the structure of language?}
\newblock In \emph{Proceedings of the 57th Annual Meeting of the Association
  for Computational Linguistics}, Florence, Italy. Association for
  Computational Linguistics.

\bibitem[{Jiang et~al.(2021)Jiang, Celikyilmaz, Smolensky, Soulos, Rao,
  Palangi, Fernandez, Smith, Bansal, and Gao}]{jiang-etal-2021-enriching}
Yichen Jiang, Asli Celikyilmaz, Paul Smolensky, Paul Soulos, Sudha Rao, Hamid
  Palangi, Roland Fernandez, Caitlin Smith, Mohit Bansal, and Jianfeng Gao.
  2021.
\newblock \href {https://aclanthology.org/2021.naacl-main.381} {Enriching
  transformers with structured tensor-product representations for abstractive
  summarization}.
\newblock In \emph{Proceedings of the 2021 Conference of the North American
  Chapter of the Association for Computational Linguistics: Human Language
  Technologies}, Online. Association for Computational Linguistics.

\bibitem[{Koupaee and Wang(2018)}]{koupaee2018wikihow}
Mahnaz Koupaee and William~Yang Wang. 2018.
\newblock \href {http://arxiv.org/abs/1810.09305} {Wikihow: A large scale text
  summarization dataset}.

\bibitem[{Kryscinski et~al.(2019)Kryscinski, Keskar, McCann, Xiong, and
  Socher}]{kryscinski-etal-2019-neural}
Wojciech Kryscinski, Nitish~Shirish Keskar, Bryan McCann, Caiming Xiong, and
  Richard Socher. 2019.
\newblock \href {https://doi.org/10.18653/v1/D19-1051} {Neural text
  summarization: A critical evaluation}.
\newblock In \emph{Proceedings of the 2019 Conference on Empirical Methods in
  Natural Language Processing and the 9th International Joint Conference on
  Natural Language Processing (EMNLP-IJCNLP)}, pages 540--551, Hong Kong,
  China. Association for Computational Linguistics.

\bibitem[{Laha et~al.(2018)Laha, Chemmengath, Agrawal, Khapra,
  Sankaranarayanan, and Ramaswamy}]{Laha2018OnCS}
Anirban Laha, Saneem~A. Chemmengath, Priyanka Agrawal, Mitesh~M. Khapra,
  Karthik Sankaranarayanan, and Harish~G. Ramaswamy. 2018.
\newblock On controllable sparse alternatives to softmax.
\newblock In \emph{Proceedings of the 32nd International Conference on Neural
  Information Processing Systems}, NIPS'18, page 6423–6433, Red Hook, NY,
  USA. Curran Associates Inc.

\bibitem[{Lavie and Agarwal(2007)}]{meteor}
Alon Lavie and Abhaya Agarwal. 2007.
\newblock Meteor: An automatic metric for mt evaluation with high levels of
  correlation with human judgments.
\newblock In \emph{Proceedings of the Second Workshop on Statistical Machine
  Translation}, StatMT '07, page 228–231, USA. Association for Computational
  Linguistics.

\bibitem[{Lester et~al.(2021)Lester, Al-Rfou, and Constant}]{Lester2021ThePO}
Brian Lester, Rami Al-Rfou, and Noah Constant. 2021.
\newblock \href {https://doi.org/10.18653/v1/2021.emnlp-main.243} {The power of
  scale for parameter-efficient prompt tuning}.
\newblock In \emph{Proceedings of the 2021 Conference on Empirical Methods in
  Natural Language Processing}, pages 3045--3059, Online and Punta Cana,
  Dominican Republic. Association for Computational Linguistics.

\bibitem[{Lewis et~al.(2020)Lewis, Liu, Goyal, Ghazvininejad, Mohamed, Levy,
  Stoyanov, and Zettlemoyer}]{lewis-etal-2020-bart}
Mike Lewis, Yinhan Liu, Naman Goyal, Marjan Ghazvininejad, Abdelrahman Mohamed,
  Omer Levy, Veselin Stoyanov, and Luke Zettlemoyer. 2020.
\newblock \href {https://doi.org/10.18653/v1/2020.acl-main.703} {{BART}:
  Denoising sequence-to-sequence pre-training for natural language generation,
  translation, and comprehension}.
\newblock In \emph{Proceedings of the 58th Annual Meeting of the Association
  for Computational Linguistics}, pages 7871--7880, Online. Association for
  Computational Linguistics.

\bibitem[{Li et~al.(2021)Li, Einolghozati, Iyer, Paranjape, Mehdad, Gupta, and
  Ghazvininejad}]{Li2021EASEES}
Haoran Li, Arash Einolghozati, Srinivasan Iyer, Bhargavi Paranjape, Yashar
  Mehdad, Sonal Gupta, and Marjan Ghazvininejad. 2021.
\newblock \href {https://doi.org/10.18653/v1/2021.newsum-1.10} {{EASE}:
  Extractive-abstractive summarization end-to-end using the information
  bottleneck principle}.
\newblock In \emph{Proceedings of the Third Workshop on New Frontiers in
  Summarization}, pages 85--95, Online and in Dominican Republic. Association
  for Computational Linguistics.

\bibitem[{Li and Hovy(2014)}]{coherence3}
Jiwei Li and Eduard~H Hovy. 2014.
\newblock A model of coherence based on distributed sentence representation.
\newblock In \emph{Proceedings of the 2014 Conference on Empirical Methods in
  Natural Language Processing}.

\bibitem[{Li and Liang(2021)}]{prefixtune}
Xiang~Lisa Li and Percy Liang. 2021.
\newblock \href {https://doi.org/10.18653/v1/2021.acl-long.353} {Prefix-tuning:
  Optimizing continuous prompts for generation}.
\newblock In \emph{Proceedings of the 59th Annual Meeting of the Association
  for Computational Linguistics and the 11th International Joint Conference on
  Natural Language Processing (Volume 1: Long Papers)}, pages 4582--4597,
  Online. Association for Computational Linguistics.

\bibitem[{Lin(2004)}]{lin-2004-rouge}
Chin-Yew Lin. 2004.
\newblock \href {https://aclanthology.org/W04-1013} {{ROUGE}: A package for
  automatic evaluation of summaries}.
\newblock In \emph{Text Summarization Branches Out}, pages 74--81, Barcelona,
  Spain. Association for Computational Linguistics.

\bibitem[{Liu et~al.(2021{\natexlab{a}})Liu, Yuan, Fu, Jiang, Hayashi, and
  Neubig}]{prompttuning_survey}
Pengfei Liu, Weizhe Yuan, Jinlan Fu, Zhengbao Jiang, Hiroaki Hayashi, and
  Graham Neubig. 2021{\natexlab{a}}.
\newblock \href {http://arxiv.org/abs/2107.13586} {Pre-train, prompt, and
  predict: {A} systematic survey of prompting methods in natural language
  processing}.
\newblock \emph{CoRR}, abs/2107.13586.

\bibitem[{Liu and Lapata(2019)}]{liu2019}
Yang Liu and Mirella Lapata. 2019.
\newblock \href {https://doi.org/10.18653/v1/P19-1500} {Hierarchical
  transformers for multi-document summarization}.
\newblock In \emph{Proceedings of the 57th Annual Meeting of the Association
  for Computational Linguistics}, pages 5070--5081, Florence, Italy.
  Association for Computational Linguistics.

\bibitem[{Liu et~al.(2021{\natexlab{b}})Liu, Zhang, Wan, Xia, He, and
  Yu}]{liu2021hetformer}
Ye~Liu, Jianguo Zhang, Yao Wan, Congying Xia, Lifang He, and Philip Yu.
  2021{\natexlab{b}}.
\newblock \href {https://doi.org/10.18653/v1/2021.emnlp-main.13} {{HETFORMER}:
  Heterogeneous transformer with sparse attention for long-text extractive
  summarization}.
\newblock In \emph{Proceedings of the 2021 Conference on Empirical Methods in
  Natural Language Processing}, pages 146--154, Online and Punta Cana,
  Dominican Republic. Association for Computational Linguistics.

\bibitem[{Manakul and Gales(2021)}]{manakul-gales-2021-long}
Potsawee Manakul and Mark Gales. 2021.
\newblock \href {https://doi.org/10.18653/v1/2021.acl-long.470} {Long-span
  summarization via local attention and content selection}.
\newblock In \emph{Proceedings of the 59th Annual Meeting of the Association
  for Computational Linguistics and the 11th International Joint Conference on
  Natural Language Processing (Volume 1: Long Papers)}, pages 6026--6041,
  Online. Association for Computational Linguistics.

\bibitem[{Mao et~al.(2021)Mao, Mathias, Hou, Almahairi, Ma, Han, tau Yih, and
  Khabsa}]{Mao2021UniPELTAU}
Yuning Mao, Lambert Mathias, Rui Hou, Amjad Almahairi, Hao Ma, Jiawei Han, Wen
  tau Yih, and Madian Khabsa. 2021.
\newblock Unipelt: A unified framework for parameter-efficient language model
  tuning.
\newblock \emph{ArXiv}, abs/2110.07577.

\bibitem[{Marcu(1997)}]{marcu-1997-discourse}
Daniel Marcu. 1997.
\newblock \href {https://aclanthology.org/W97-0713} {From discourse structures
  to text summaries}.
\newblock In \emph{Intelligent Scalable Text Summarization}.

\bibitem[{Martins et~al.(2020)Martins, Marinho, and
  Martins}]{Martins2020SparseTG}
Pedro~Henrique Martins, Zita Marinho, and Andr{\'e} F.~T. Martins. 2020.
\newblock \href {https://doi.org/10.18653/v1/2020.emnlp-main.348} {Sparse text
  generation}.
\newblock In \emph{Proceedings of the 2020 Conference on Empirical Methods in
  Natural Language Processing (EMNLP)}, pages 4252--4273, Online. Association
  for Computational Linguistics.

\bibitem[{Nan et~al.(2021)Nan, Radev, Zhang, Rau, Sivaprasad, Hsieh, Tang,
  Vyas, Verma, Krishna, Liu, Irwanto, Pan, Rahman, Zaidi, Mutuma, Tarabar,
  Gupta, Yu, Tan, Lin, Xiong, Socher, and Rajani}]{nan2021dart}
Linyong Nan, Dragomir Radev, Rui Zhang, Amrit Rau, Abhinand Sivaprasad,
  Chiachun Hsieh, Xiangru Tang, Aadit Vyas, Neha Verma, Pranav Krishna,
  Yangxiaokang Liu, Nadia Irwanto, Jessica Pan, Faiaz Rahman, Ahmad Zaidi,
  Murori Mutuma, Yasin Tarabar, Ankit Gupta, Tao Yu, Yi~Chern Tan, Xi~Victoria
  Lin, Caiming Xiong, Richard Socher, and Nazneen~Fatema Rajani. 2021.
\newblock Dart: Open-domain structured data record to text generation.
\newblock \emph{arXiv preprint arXiv:2007.02871}.

\bibitem[{Narayan et~al.(2018)Narayan, Cohen, and
  Lapata}]{narayan-etal-2018-dont}
Shashi Narayan, Shay~B. Cohen, and Mirella Lapata. 2018.
\newblock \href {https://doi.org/10.18653/v1/D18-1206} {Don{'}t give me the
  details, just the summary! topic-aware convolutional neural networks for
  extreme summarization}.
\newblock In \emph{Proceedings of the 2018 Conference on Empirical Methods in
  Natural Language Processing}, pages 1797--1807, Brussels, Belgium.
  Association for Computational Linguistics.

\bibitem[{Novikova et~al.(2017)Novikova, Du{\v{s}}ek, and
  Rieser}]{novikova2017e2e}
Jekaterina Novikova, Ondrej Du{\v{s}}ek, and Verena Rieser. 2017.
\newblock \href {https://arxiv.org/abs/1706.09254} {The {E2E} dataset: New
  challenges for end-to-end generation}.
\newblock In \emph{Proceedings of the 18th Annual Meeting of the Special
  Interest Group on Discourse and Dialogue}, Saarbr\"ucken, Germany.
\newblock ArXiv:1706.09254.

\bibitem[{Papineni et~al.(2002)Papineni, Roukos, Ward, and Zhu}]{bleu}
Kishore Papineni, Salim Roukos, Todd Ward, and Wei-Jing Zhu. 2002.
\newblock \href {https://doi.org/10.3115/1073083.1073135} {Bleu: A method for
  automatic evaluation of machine translation}.
\newblock In \emph{Proceedings of the 40th Annual Meeting on Association for
  Computational Linguistics}, ACL '02, page 311–318, USA. Association for
  Computational Linguistics.

\bibitem[{Pasunuru et~al.(2021)Pasunuru, Celikyilmaz, Galley, Xiong, Zhang,
  Bansal, and Gao}]{pasunuru2021data}
Ramakanth Pasunuru, Asli Celikyilmaz, Michel Galley, Chenyan Xiong, Yizhe
  Zhang, Mohit Bansal, and Jianfeng Gao. 2021.
\newblock \href {https://arxiv.org/pdf/2103.01863.pdf} {Data augmentation for
  abstractive query-focused multi-document summarization}.
\newblock In \emph{AAAI}.

\bibitem[{Petroni et~al.(2019)Petroni, Rockt{\"a}schel, Riedel, Lewis, Bakhtin,
  Wu, and Miller}]{petroni-etal-2019-language}
Fabio Petroni, Tim Rockt{\"a}schel, Sebastian Riedel, Patrick Lewis, Anton
  Bakhtin, Yuxiang Wu, and Alexander Miller. 2019.
\newblock \href {https://doi.org/10.18653/v1/D19-1250} {Language models as
  knowledge bases?}
\newblock In \emph{Proceedings of the 2019 Conference on Empirical Methods in
  Natural Language Processing and the 9th International Joint Conference on
  Natural Language Processing (EMNLP-IJCNLP)}, pages 2463--2473, Hong Kong,
  China. Association for Computational Linguistics.

\bibitem[{PurdueOWL(2019)}]{news}
PurdueOWL. 2019.
\newblock Journalism and journalistic writing: The inverted pyramid structure.

\bibitem[{Radford et~al.(2019)Radford, Wu, Child, Luan, Amodei, and
  Sutskever}]{radford2019language}
Alec Radford, Jeff Wu, Rewon Child, David Luan, Dario Amodei, and Ilya
  Sutskever. 2019.
\newblock Language models are unsupervised multitask learners.

\bibitem[{Raffel et~al.(2020)Raffel, Shazeer, Roberts, Lee, Narang, Matena,
  Zhou, Li, and Liu}]{2020t5}
Colin Raffel, Noam Shazeer, Adam Roberts, Katherine Lee, Sharan Narang, Michael
  Matena, Yanqi Zhou, Wei Li, and Peter~J. Liu. 2020.
\newblock \href {http://jmlr.org/papers/v21/20-074.html} {Exploring the limits
  of transfer learning with a unified text-to-text transformer}.
\newblock \emph{Journal of Machine Learning Research}, 21(140):1--67.

\bibitem[{Roy et~al.(2021)Roy, Saffar, Vaswani, and
  Grangier}]{10.1162/tacl_a_00353}
Aurko Roy, Mohammad Saffar, Ashish Vaswani, and David Grangier. 2021.
\newblock \href {https://doi.org/10.1162/tacl_a_00353} {{Efficient
  Content-Based Sparse Attention with Routing Transformers}}.
\newblock \emph{Transactions of the Association for Computational Linguistics},
  9:53--68.

\bibitem[{See et~al.(2017)See, Liu, and Manning}]{See2017GetTT}
Abigail See, Peter~J. Liu, and Christopher~D. Manning. 2017.
\newblock \href {https://doi.org/10.18653/v1/P17-1099} {Get to the point:
  Summarization with pointer-generator networks}.
\newblock In \emph{Proceedings of the 55th Annual Meeting of the Association
  for Computational Linguistics (Volume 1: Long Papers)}, pages 1073--1083,
  Vancouver, Canada. Association for Computational Linguistics.

\bibitem[{Shi et~al.(2021)Shi, Gao, Ren, Xu, Liang, Li, and
  Kwok}]{Shi2021SparseBERTRT}
Han Shi, Jiahui Gao, Xiaozhe Ren, Hang Xu, Xiaodan Liang, Zhenguo Li, and James
  Tin-Yau Kwok. 2021.
\newblock \href {https://proceedings.mlr.press/v139/shi21a.html} {Sparsebert:
  Rethinking the importance analysis in self-attention}.
\newblock In \emph{Proceedings of the 38th International Conference on Machine
  Learning}, volume 139 of \emph{Proceedings of Machine Learning Research},
  pages 9547--9557. PMLR.

\bibitem[{Smith et~al.(2022)Smith, Patwary, Norick, LeGresley, Rajbhandari,
  Casper, Liu, Prabhumoye, Zerveas, Korthikanti, Zheng, Child, Aminabadi,
  Bernauer, Song, Shoeybi, He, Houston, Tiwary, and Catanzaro}]{megatron}
Shaden Smith, Mostofa Patwary, Brandon Norick, Patrick LeGresley, Samyam
  Rajbhandari, Jared Casper, Zhun Liu, Shrimai Prabhumoye, George Zerveas,
  Vijay Korthikanti, Elton Zheng, Rewon Child, Reza~Yazdani Aminabadi, Julie
  Bernauer, Xia Song, Mohammad Shoeybi, Yuxiong He, Michael Houston, Saurabh
  Tiwary, and Bryan Catanzaro. 2022.
\newblock \href {http://arxiv.org/abs/2201.11990} {Using deepspeed and megatron
  to train megatron-turing {NLG} 530b, {A} large-scale generative language
  model}.
\newblock volume abs/2201.11990.

\bibitem[{Sukhbaatar et~al.(2021)Sukhbaatar, Ju, Poff, Roller, Szlam, Weston,
  and Fan}]{expirespan}
Sainbayar Sukhbaatar, Da~Ju, Spencer Poff, Stephen Roller, Arthur Szlam, Jason
  Weston, and Angela Fan. 2021.
\newblock \href {https://arxiv.org/pdf/2105.06548.pdf} {Not all memories are
  created equal: Learning to forget by expiring}.
\newblock In \emph{Proceedings of the 37th International Conference on Machine
  Learning, ICML 2020, 13-18 July 2020, Virtual Event}.

\bibitem[{Sun and Lu(2020)}]{sun-lu-2020-understanding}
Xiaobing Sun and Wei Lu. 2020.
\newblock \href {https://doi.org/10.18653/v1/2020.acl-main.312} {Understanding
  attention for text classification}.
\newblock In \emph{Proceedings of the 58th Annual Meeting of the Association
  for Computational Linguistics}, pages 3418--3428, Online. Association for
  Computational Linguistics.

\bibitem[{Tay et~al.(2021)Tay, Dehghani, Abnar, Shen, Bahri, Pham, Rao, Yang,
  Ruder, and Metzler}]{tay2021long}
Yi~Tay, Mostafa Dehghani, Samira Abnar, Yikang Shen, Dara Bahri, Philip Pham,
  Jinfeng Rao, Liu Yang, Sebastian Ruder, and Donald Metzler. 2021.
\newblock \href {https://openreview.net/forum?id=qVyeW-grC2k} {Long range arena
  : A benchmark for efficient transformers}.
\newblock In \emph{International Conference on Learning Representations}.

\bibitem[{Wang et~al.(2020)Wang, Li, Khabsa, Fang, and Ma}]{linformer}
Sinong Wang, Belinda~Z. Li, Madian Khabsa, Han Fang, and Hao Ma. 2020.
\newblock \href {http://arxiv.org/abs/2006.04768} {Linformer: Self-attention
  with linear complexity}.
\newblock \emph{CoRR}, abs/2006.04768.

\bibitem[{Welleck et~al.(2020)Welleck, Kulikov, Roller, Dinan, Cho, and
  Weston}]{unlikelihood}
Sean Welleck, Ilia Kulikov, Stephen Roller, Emily Dinan, Kyunghyun Cho, and
  Jason Weston. 2020.
\newblock \href {https://openreview.net/forum?id=SJeYe0NtvH} {Neural text
  generation with unlikelihood training}.
\newblock In \emph{International Conference on Learning Representations}.

\bibitem[{Xiao et~al.(2021)Xiao, Beltagy, Carenini, and Cohan}]{xiao2021primer}
Wen Xiao, Iz~Beltagy, Giuseppe Carenini, and Arman Cohan. 2021.
\newblock \href {http://arxiv.org/abs/2110.08499} {Primer: Pyramid-based masked
  sentence pre-training for multi-document summarization}.

\bibitem[{Yates et~al.(2021)Yates, Nogueira, and Lin}]{beyondbert}
Andrew Yates, Rodrigo Nogueira, and Jimmy Lin. 2021.
\newblock \href {https://doi.org/10.1145/3437963.3441667} {Pretrained
  transformers for text ranking: Bert and beyond}.
\newblock In \emph{Proceedings of the 14th ACM International Conference on Web
  Search and Data Mining}, WSDM '21, page 1154–1156, New York, NY, USA.
  Association for Computing Machinery.

\bibitem[{Zaheer et~al.(2020)Zaheer, Guruganesh, Dubey, Ainslie, Alberti,
  Ontanon, Pham, Ravula, Wang, Yang et~al.}]{bigbird}
Manzil Zaheer, Guru Guruganesh, Kumar~Avinava Dubey, Joshua Ainslie, Chris
  Alberti, Santiago Ontanon, Philip Pham, Anirudh Ravula, Qifan Wang, Li~Yang,
  et~al. 2020.
\newblock \href {https://arxiv.org/pdf/2007.14062.pdf} {Big bird: Transformers
  for longer sequences}.
\newblock volume~33.

\bibitem[{Zhang et~al.(2020)Zhang, Zhao, Saleh, and Liu}]{pegasus}
Jingqing Zhang, Yao Zhao, Mohammad Saleh, and Peter~J. Liu. 2020.
\newblock \href {http://proceedings.mlr.press/v119/zhang20ae.html} {Pegasus:
  Pre-training with extracted gap-sentences for abstractive summarization}.
\newblock In \emph{Proceedings of the 37th International Conference on Machine
  Learning, ICML 2020, 13-18 July 2020, Virtual Event}, volume 119 of
  \emph{Proceedings of Machine Learning Research}, pages 11328--11339. PMLR.

\bibitem[{Zhang* et~al.(2020)Zhang*, Kishore*, Wu*, Weinberger, and
  Artzi}]{bertscore}
Tianyi Zhang*, Varsha Kishore*, Felix Wu*, Kilian~Q. Weinberger, and Yoav
  Artzi. 2020.
\newblock \href {https://openreview.net/forum?id=SkeHuCVFDr} {Bertscore:
  Evaluating text generation with bert}.
\newblock In \emph{International Conference on Learning Representations}.

\bibitem[{Zhao et~al.(2019)Zhao, Peyrard, Liu, Gao, Meyer, and
  Eger}]{moverscore}
Wei Zhao, Maxime Peyrard, Fei Liu, Yang Gao, Christian~M. Meyer, and Steffen
  Eger. 2019.
\newblock \href {https://doi.org/10.18653/v1/D19-1053} {{M}over{S}core: Text
  generation evaluating with contextualized embeddings and earth mover
  distance}.
\newblock In \emph{Proceedings of the 2019 Conference on Empirical Methods in
  Natural Language Processing and the 9th International Joint Conference on
  Natural Language Processing (EMNLP-IJCNLP)}, pages 563--578, Hong Kong,
  China. Association for Computational Linguistics.

\end{thebibliography}
\bibliographystyle{acl_natbib}

\newpage
\appendix

\section{Appendix}

\begin{figure}[htbp]
    \centering
    \begin{tcolorbox}[boxsep=0pt,left=0pt,colframe=white]
    \footnotesize
    \begin{tabular}{p{\linewidth}}
    \textbf{Context}: How to straighten a curly wig?\\
    \hline
    It is the best method for both straightening and curling a wig. It is also the best method for synthetic wigs that may be more delicate to heated styling tools. \\ Allow the hair to fall straight down without resting on anything. Tripods work brilliantly if you fit a mannequin head on them. \\ Boil about 1.5L for a long wig and 1L or less for a short wig. The heat of the water will soften the plastic fibers and will pull them down. Use a wide soft-bristled brush. \\ For extremely curly wigs, it will take more than one treatment and a lot of time. It does no damage to the wig and can be done while the wig is still wet. This is a permanent style; you will need to re-curl the wig if you don't want it straight anymore. You're basically remolding plastic into a new shape, so will need to mold it back if you want the curls again.\\
    \hline
    \textbf{Output Summaries}:\\
    \hline
    \textbf{Human }: consider trying the hot water method. place the wig on a stand. boil a pot full of water. pour it carefully over the wig, making sure that it completely covers it. avoid combing or brushing a wig when it is wet. wait until it is dry. repeat the method if necessary. \\
    \textbf{Finetune}: use this method \textcolor{red}{for very curly wigs}. place the wig on a flat surface. \textcolor{red}{boil water} in a large pot. remove the pot from the heat and \textcolor{red}{allow the water to cool}. brush the wig with a wide-bristled brush. repeat the process until the entire wig is straight. curl the wig if you don't want it straight anymore. \\
    \textbf{Prefixtune}:  understand the benefits of this method. lay the wig flat on a flat surface. \textcolor{red}{boil a small amount of water}. brush the wig with a soft-bristled brush. repeat the process if necessary. re-curl the wig if desired. finished. \\
    \textbf{HSoftSA}: use this method if you want to straighten the wig. place the wig on a mannequin head. boil a pot of water. brush the wig with a soft-bristled brush. repeat as needed. re-curl the wig if necessary. \\
    \textbf{HierBlock}: heat the water in a large pot over medium heat. put the wig in the pot and allow it to sit for a few minutes. remove the wig from the pot. brush the wig with a soft-bristled brush. repeat the process with the other wig.\\
    \textbf{HierBlock+SoftSA}: wash the wig with \textcolor{red}{warm} water. put the wig on a mannequin head. rinse the wig. brush the hair with a soft-bristled brush. repeat the process until the hair is completely straight. re-curl the wig if you want. \\
    \end{tabular}
    \end{tcolorbox}    
    \vspace{-0.3cm}
    \caption{Model Generated Output Text on Wikihow Dataset. The red colored text indicates factual errors, repetitions, and incoherent text.}  
    \label{fig:generationswikihow}
\end{figure}

\begin{table*}[t]
    \centering
    \small
        \begin{tabular}{l@{}ccccccc}
        \hline 
        \textbf{Parameter} & \textbf{Xsum} & \textbf{CNN/DM} & \textbf{PubMed} & \textbf{Wikihow} & \textbf{SAMSum} &\textbf{E2E} & \textbf{DART} \\
        \hline
        learning rate & 5e-05&5e-05 & 5e-05& 5e-05& 5e-05& 5e-05& 5e-05\\
        $\#$ epochs & 30 & 30 & 30 & 30 & 20 & 10 & 10 \\
        batch size & 8 & 8 & 8 & 8 & 16 & 16 & 16\\
        prefix-length & 100 & 50-100 & 100-200 & 100-200 & 10-20 & 5-10 & 5-10\\
        beamsize & 5 & 5 & 5 & 5 & 5 & 5 & 5\\
        \hline 
        Hierarchical Blocking & & & & & & & \\
        \hline
        encoder block size & 1,2,3 & 1,2,3& 1,2,3& 1,2,3& 1,2,3& 1,2,3& 1,2,3\\
        decoder block size & 1,2& 1,2& 1,2& 1,2& 1,2& 1,2& 1,2\\
        \hline
        Sparse Attention & & & & & & & \\
        \hline
        top-\textit{p} & 95.\%& 95.\%& 95.\%& 95.\%& 95.\%& 95.\%& 95.\%\\
        $\tau$ (top-\textit{p}) & 1.0,0.1& 1.0,0.1& 1.0,0.1& 1.0,0.1& 1.0,0.1& 1.0,0.1& 1.0,0.1\\
        $\tau$ (soft attn.) & 1.0,0.1,0.01 & 1.0,0.1,0.01& 1.0,0.1,0.01& 1.0,0.1,0.01& 1.0,0.1,0.01& 1.0,0.1,0.01& 1.0,0.1,0.01\\
        \hline
    \end{tabular}
    \caption{Hyperparameters of different prefix-tuned models.}
    \label{apptable:hyperparameters}
\end{table*}

\begin{table*}[t]
    \centering
    \small
        \begin{tabular}{lllll}
        \hline
        Corpus & Version & License & Citation & Link \\
        \hline
        XSum & v1 & MIT & \citet{narayan-etal-2018-dont} & \tiny{\url{https://github.com/EdinburghNLP/XSum}} \\
        CNN/DM & v1 & MIT & \citet{hermann2015teaching} & \tiny{\url{https://github.com/abisee/cnn-dailymail}}\\
        PubMed & v1 &Creative Commons & \citet{cohan-etal-2018-discourse} & \tiny{\url{https://github.com/armancohan/long-summarization}}\\
        WikiHow & v1 & CC-BY-NC-SA & \citet{koupaee2018wikihow} & \tiny{\url{https://github.com/mahnazkoupaee/WikiHow-Dataset}}\\
        SAMSum & v1 & CC BY-NC-ND 4.0 & \citet{gliwa2019samsum} & \tiny{\url{https://github.com/giancolu/Samsung-dataset}} \\
        E2E &v1 & CC4.0-BY-SA & \citet{e2e} & \tiny{\url{https://github.com/tuetschek/e2e-cleaning}}\\
        DART & v1 & MIT &  \citet{nan2021dart} & \tiny{\url{https://github.com/Yale-LILY/dart}} \\
        \hline
        \end{tabular}
\caption{Additional documentation of scientific artifacts used in our paper.}
\label{app:licences}
\end{table*}

\subsection{Hyperparameters (Cont. from $\S$~\ref{setup})}
\label{app:hyperparameters}

We fit our BART-LARGE models to their respective datasets 
with the hyperparameters shown in Table~\ref{apptable:hyperparameters}. 
Encoder/decoder block sizes indicate the size of the segments we split the input/output tokens. For instance, if the encoder block size is 2, we split the input tokens into two segments. Each segment has designated set of prefixes which can vary at each layer. In hierarchical blocking models (HierBlock) we segment the lower layers, so the prefixes are blocked for different segments, while at the top layers no segmentation or blocking is applied. We use at most two segments in the output text since the text generations tasks we investigate in this work contain much shorter output tokens compared to the input tokens.  

\textbf{Gumbel Softmax Reparameterization Trick}:
Sampling introduces discrete valued parameters which are not differential at training time. 
Thus, we resort to the GumbelSoftmax trick \citep{jang2017categorical}, which provides a
tool for sampling from a continuous approximation of a discrete distribution.
The Gumbel-Softmax trick considers a discrete variable with class probabilities $\pi_1, \cdots, \pi_k$, and draws samples $g_1, \cdots, g_k$
from a Gumbel distribution, Gumbel(0,1), as follows:
\begin{equation}
y = \frac{\exp(( \log(\pi_i) + g_i)/\tau )}
{\sum^k_{j=1} \exp((\log \pi_j +g_j )/\tau )}
\label{g}
\end{equation}

for $i=1$,$\dots,k$. The Gumbel(0,1) distribution can be sampled using inverse transform sampling by drawing $u \sim$ Uniform(0,1) and computing $g=-\log(-\log(u))$. Plugging in the samples $g_i$ and the class probabilities $\pi_i$ in Eq.~\ref{g}, we generate a $k$-dimensional sample
vector $y\in$$\triangle^{k-1}$, that is the continuous approximation of
the one-hot-encoded representation of the discrete
variable $d$. In fact, as $\tau$ approaches 0, samples from
the Gumbel-Softmax distribution become one-hot,
making it discrete. For more details please refer to \citep{jang2017categorical}.

\subsection{Dataset Details (Cont. from $\S$\ref{datasetsintext})}

\label{app:dataset_details}

All datasets are in English language. The summarization datasets range from extreme abstractive summarization with XSum \cite{narayan-etal-2018-dont} to summarize documents into one summary sentence, conversational summarization using SAMSum dataset \citep{gliwa2019samsum}, long clinical document summarization with PubMed \citep{cohan-etal-2018-discourse}\footnote{We acknowledge that the source of dataset is the NLM Catalog, and the citations used in Pubmed corpus may not reflect the most current/accurate data available from NLM, which is updated regularly.}
and DIY domain with Wikihow \citep{koupaee2018wikihow}, and commonly used CNN/DM \citep{hermann2015teaching,See2017GetTT} news article summarization dataset with an ”Inverted Pyramid” \citep{news} document structure \citep{kryscinski-etal-2019-neural}. We also investigate S2T datasets on customer reviewers including E2E \citep{novikova2017e2e,e2e} and DART \citep{nan2021dart} with each input being a semantic RDF triple set  derived from data records in tables and sentence descriptions that cover all facts in the triple set.

\paragraph{XSum} \citep{narayan-etal-2018-dont} is a collection of ~227k BBC News articles ranging from 2010 to 2017. The dataset covers a wide range of subjects. The single-sentence summaries are written by professionals.

\paragraph{CNN/DailyMail} \citep{hermann2015teaching} dataset contains ~93k news articles extracted from CNN News, and around 220k articles extracted from the Daily Mail
newspapers. The summaries are human written bullet point text which are provided in the same source documents. In our experiments we use the non-anonymized version, which is commonly used in summarization research papers.

\paragraph{PubMed} \citep{cohan-etal-2018-discourse} is a  long document dataset of ~215K scientific publications from
PubMed. The task is to generate the abstract from the paper body.

\paragraph{WikiHow} \citep{koupaee2018wikihow}  is a large-scale dataset of ~200K instructions from the online WikiHow.com website. Each instance consists of multiple instruction-step paragraphs and an accompanying summary sentence of each paragraph. The task is
to generate the concatenated summary-sentences from the
paragraphs.

\paragraph{SAMSum} \citep{gliwa2019samsum} is a multi-turn dialog corpus of 16K chat dialogues and manually annotated summaries.  The task is to generate an abstractive summary of the dialog with coherent discourse structure of the original dialog. 

\paragraph{E2E} \citep{e2e} is a structured data to natural langauge summary dataset that provides information about restaurants. The structured inputs consists of different attributes (slots) such as name, type of food or area and their values.   It contains ~50K instances of diverse descriptions of the structured input introducing challenges, such as open vocabulary,
complex syntactic structures and diverse discourse
phenomena.

\paragraph{DART}  \citep{nan2021dart} is a text generation dataset for open-domain structured data-record to text generation. It  consists of ~82K examples from variety of domains. The inputs are in semantic RDF triple set  form which are derived from data records in tables and tree ontology of the schema. The output generations are human annotated with sentence descriptions that cover all facts in the triple set.

\paragraph{Licence details} In our experiments, we use several datasets (as detailed above) from public resources . Table~\ref{app:licences} summarizes the licences. All data are solely used for research purposes.

\subsection{Compute Infrastructure and Run time}
Each experiment runs on a single machine with 
8 GPUs. Depending on the training dataset size, summarization models require from 5.5 hours to 18 hours to train. The data-to-text datasets are much smaller which takes less than 4 hours. 
All fine-tuned models follow the BART-large transformer architecture with a total of 12 layers, 1024 hidden dimensions, and 406M parameters. The prefix-models increase the parameters size of fine-tune models by 0.1\% up to 2\% depending on the number of prefix parameters. See hyperparameters details in Appendix~\ref{app:hyperparameters}.

\subsection{Visualization of Prefix Parameters (Cont. from $\S$~\ref{discourse}}
\label{app:prefixvisual}
\begin{figure*}[th!]
    \centering
    \adjustbox{trim={.1\width} {.1\height} {0.1\width} {.1\height},clip}
    {\includegraphics[scale=0.38]{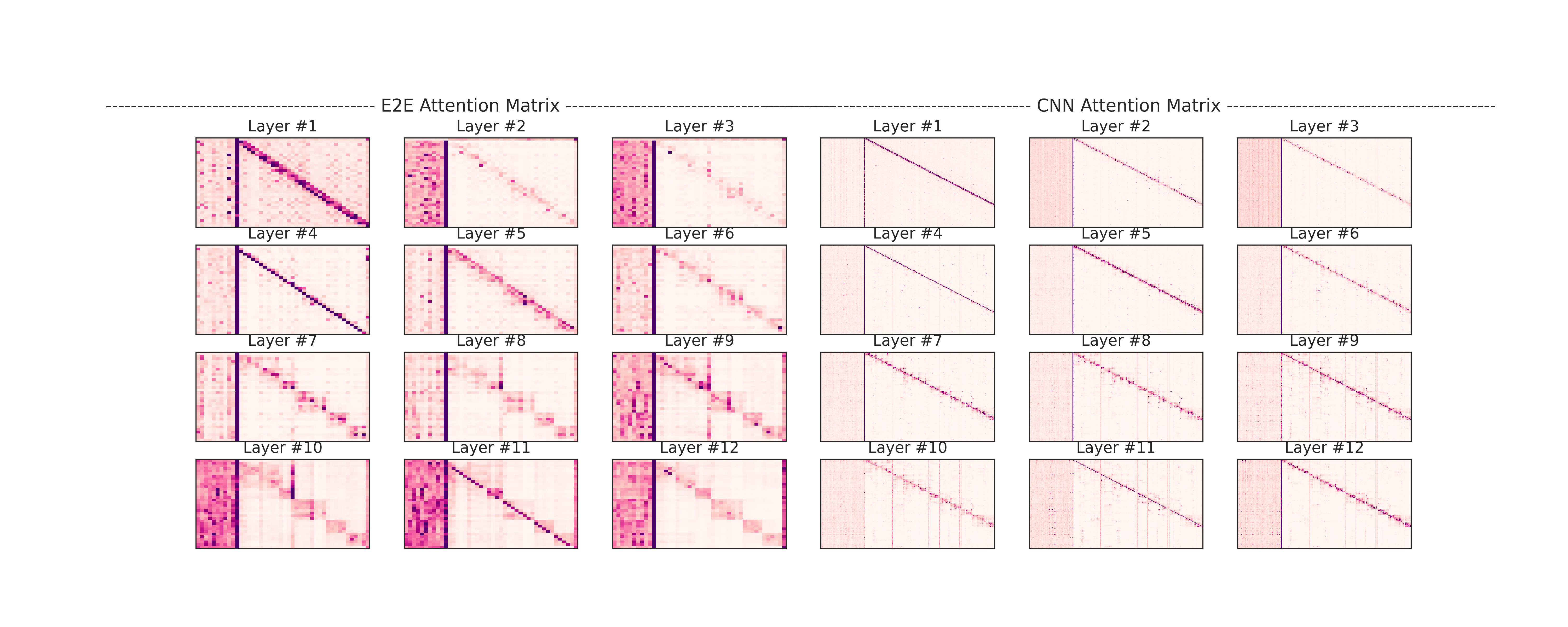} }
    \caption{\small Encoder self-attention matrices \textbf{\textit{A}} of  prefix-tuned models indicating the query attention scores over all keys (prefix+inputs) on the y-axis. The scores are averaged over all heads. The left block is for E2E dataset where the first 10 features represent prefix features, while CNN/DM dataset on the right with first 100 features represent the prefixes. }%
    \label{app:fig:attn}%
\end{figure*}
To analyze the attention behaviour (similar to \citep{sun-lu-2020-understanding}) we plot the attention matrix of the prefix-tuned models focusing on the prefix parameters. We use a prefix-tuned BART-LARGE (12-layer stacked transformer) on two tasks: data-to-text generation on E2E \citep{e2e} and summarization on CNN/DM \citep{hermann2015teaching}. In Figure~\ref{app:fig:attn}, we plot the encoder self-attention distributions \textbf{\textit{A}} for different layers averaging over head vectors. The \textit{x}-axis represent the keys, while y-axis denote the queries.



\begin{table*}[t!]
    \centering
    \footnotesize
        \begin{tabular}{@{}lcccccc|ccccc@{}}
        \hline 
        &  \multicolumn{5}{c}{\textbf{E2E}} & \multicolumn{5}{c}{\textbf{DART}} \\
        \textbf{Method} & \#Parm. & BLEU & NIST & MET & R-L & CIDEr & BLEU & MET & TER ↓ & Mover & BERT \\
        \hline
        \multicolumn{12}{l}{\textbf{GPT-2-Large}}  \\
        \hline
        Finetune (*) & 774M  & 
        68.5 & 8.78 &46.0& 69.9 &2.45 & 
        \textbf{47.0} & \textbf{0.39} & \textbf{0.46} & 0.51 & 0.94 \\
        
        Prefixtune (*) & 774M+\%0.1  & 
        \textbf{70.3} &\textbf{8.85}& \textbf{46.2}& \textbf{71.7} &\textbf{2.47}  & 
        46.7 &\textbf{0.39} & 0.45 & 0.51 & 0.94 \\
        \hline
        \multicolumn{12}{l}{\textbf{BART-Large}}  \\
        \hline
        Finetune & 406M & \textbf{67.8} & \textbf{8.76} & \textbf{45.1}& \textbf{69.5} &\textbf{2.38} & 
        46.1 & \underline{\textbf{0.38}} & \textbf{0.46} & \textbf{\underline{0.53}} & \textbf{0.95}  \\
        Prefixtune & 406M+\%0.1   &  \textbf{\underline{67.3}} & 8.66 & 44.8& 68.6 & 2.34 &
        45.9 & \textbf{0.39} & \underline{\textbf{0.45}} & \underline{\textbf{0.53}} & \textbf{0.95} \\
        Uniblock  &  406M+\%0.1 & 66.1 & 8.60 & \textbf{\underline{45.0}} & 68.3 & \textbf{\underline{2.36}} & 
        \underline{\textbf{46.5}} & \textbf{0.39} & \textbf{0.46}  & \textbf{\underline{0.53}} & \textbf{0.95}  \\
        HierBlock & 406M+\%0.1& 67.2 & \textbf{\underline{8.70}}      & \textbf{45.1} & \textbf{\underline{69.1}} & 2.35 & \textbf{46.6} & \textbf{0.39} & \textbf{\underline{0.45}} & \textbf{0.54} & \textbf{0.95}  \\
        \hline
    \end{tabular}
    \caption{\small \textbf{Data-to-Text}: \textbf{Prefix-Blocking}  Models compared against Finetune and Prefixtune models. (\textbf{*}) Top block are best reported numbers in  \cite{prefixtune} using GPT-2 LARGE model, twice the size of BART-LARGE. Bottom block are our experiment results. Best results on GPT-2 and BART-LARGE models are \textbf{bolded}  and second best BART-LARGE models are also \textbf{\underline{underlined}}. For all the prefix-tuned BART models, the prefix-length is 5 for E2E and 10 for DART dataset. All hyper-parameters are summarized in Appendix Table~\ref{apptable:hyperparameters}.}
    \label{apptab:table2textblockingonly}
\end{table*}

\subsection{Are All Prefix Parameters Useful? (Cont. from $\S$~\ref{exp:prefixanalysis})} 
\label{app:layerwiseprefix}
We investigate the influence of prefix parameters on different layers of the network. For this experiments we trained BART-LARGE and introduced prefix parameters only at the top layers, lower layers and all layers (this is same as baseline prefix-tuning models). On XSum dataset, we observed 
a large performance gap between
the models trained with top/lower layers, 
while we obtain the best performance
when we tune all-layer prefix parameters (in Table~\ref{tab:minirouge} in the main text).
Here, we investigate if similar perforamance gains are observed on 
dialog summarization (SAMSum) and data to text generation (E2E) tasks.

We show the performance scores of our experiments on 
validation datasets in Table~\ref{tab:appres4}. We observe similar results as the analysis on XSum dataset. Top layers prefix parameters learn salient features related to the task, though using prefixes at all layers yields better performance.

\begin{table}[h]
    \centering
    \footnotesize
    \begin{tabular}{lcc}
        \hline 
        & \textbf{SAMSum} & \textbf{E2E} \\
        \textbf{Method} & R1/R2/RL & BLEU/RL \\
         \hline
          Top (8-12)     & 49.55/23.72/42.16 & 64.3/65.7 \\
          Low (1-7)   &  43.16/19.08/38.14 &  62.4/62.6\\
         \hline
        
         All (1-12)   & 50.16/25.03/43.16 & 65.4/66.5\\
        \hline
    \end{tabular}
    \vspace{-0.3cm}
    \caption{Results of prefix-
tuned models on validation datasets of SAMSum (from the summarization task) and E2E (from the structure to text task) using only the top/low layers.}
    \label{tab:appres4}
    \vspace{-0.3cm}
\end{table}

\subsection{Investigation of Hierarchical Prompt Design (Cont. from $\S$~\ref{blockingresults})}
\label{app:hierpromptdesign}
We investigate if blocking prefixes helps for data-to-text tasks. Table~\ref{apptab:table2textblockingonly} shows the results.  Similar to summarization experiments in $\S$~\ref{blockingresults}, we observe improvements with hierarchical blocking on E2E dataset, though the improvement is minimal in DART dataset. We find that the structural bias we apply to the prefix-tuning models using hierarchical blocking is more prominent in summarization tasks and we see better improvements over prefix-tuning.
We also include previous best model results reported in \citep{prefixtune}. Their results are from GPT-2 LARGE, which is twice the size of BART-LARGE models, so our results are slighly lower. We replicated the fine-tuning and prefixtuning results for fair comparison. We provide the model sizes in terms of number of parameters in Table~\ref{apptab:table2textblockingonly}.
We conclude from these results that the prefix models tuned with structurally biased additional set of parameters can yield more robust information transfer reaching as good as finetuning models.   
In Figure~\ref{fig:generationswikihow} we show the output summaries generated by some of our best discourse aware prefix-tuned models in comparison to baseline fine-tuned and prefix-tuned models.



\begin{table*}[t!]
    \centering
    \footnotesize
        \begin{tabular}{@{}l@{}ccccc}
        \hline 
        & \textbf{Xsum} & \textbf{CNN/DM} & \textbf{PubMed} & \textbf{Wikihow} & \textbf{SAMSum} \\
        \textbf{Method} \ \ \ & R1/R2/R-L & R1/R2/R-L & R1/R2/R-L & R1/R2/R-L & R1/R2/R-L\\
        \hline
        \multicolumn{6}{l}{\textit{\textbf{Finetune}}} \\
        \hline
        Dense (reproduce) \ \ \ \  & 
        \textbf{45.47}/\textbf{22.40}/\textbf{37.42} &
        \textbf{44.30}/\textbf{21.14}/\textbf{41.22} & 
        \textbf{45.09}/\textbf{19.56}/27.42 &  
        43.26/19.38/34.89 &
        \textbf{53.02}/\textbf{28.30}/\textbf{48.73} \\
        (\textit{a}) TruncSA   & 
        45.17/21.98/37.02& 
        43.02/20.03/39.96& 
        41.08/17.02/26.37& 
        \textbf{44.85}/\textbf{20.00}/\textbf{35.58}& 
        52.45/27.56/48.12
        \\
        (\textit{b}) SoftSA   & 
        45.34/22.02/37.23& 
        42.97/20.44/39.67 & 
        40.15/16.30/26.26& 
        41.56/17.80/32.53&
        52.47/27.88/48.45
        \\
        \hline
        \multicolumn{6}{l}{\textit{\textbf{Prefix-tune}}} \\
        \hline
        Dense (reproduce)   & 
        43.43/20.37/35.47 & 
        42.50/19.52/39.08 & 
        42.38/16.31/24.57 & 
        39.26/15.58/28.27 &
        52.12/26.52/48.05
        \\
        (\textit{a}) TruncSA    &
        43.56/20.62/35.96& 
        42.80/19.81/39.52 & 
        42.50/16.85/\textbf{\underline{25.61}} & 
        39.43/15.69/28.94 &
        52.09/27.49/47.88\\
        (\textit{b}) SoftSA    & 
        43.80/20.82/35.94& 
        42.25/19.50/39.24& 
        & 
        39.31/15.61/28.94&
        51.72/25.51/47.35\\
        (\textit{c}) HTruncSA \ \ \     & 
        \textbf{\underline{44.17}}/\textbf{\underline{21.11}}/\textbf{\underline{36.42}}& 
        \textbf{\underline{43.17}}/\textbf{\underline{20.17}}/\textbf{\underline{40.00}}& 
        \textbf{\underline{43.30}}/\textbf{\underline{17.00}}/25.28& 
        \textbf{\underline{40.00}}/\textbf{\underline{16.11}}/\textbf{\underline{30.02}} &
        52.12/26.94/48.00
        \\
        (\textit{d}) HSoftSA     & 
        44.05/20.84/36.13 & 
        43.10/20.06/39.83& 
        42.30/16.20/24.90& 
        39.83/16.10/30.01 &
        \textbf{\underline{52.37}}/\textbf{\underline{27.57}}/\textbf{\underline{48.33}}\\
        \hline
    \end{tabular}
    \caption{\small \textbf{Summarization}: \textbf{Sparse Attention} experiment results on Finetuning, and Prefix-tuning with Truncated (TruncSA) and Bernoulli Sampling soft attention (SoftSA) and Hierarchical Truncated (HTruncSA) and Soft Attention (HSoftSA) for Prefix-Tuning. All models are based on BART-LARGE. Best finetune results accross models are \textbf{bolded} while the best prefixtune models are also \textbf{\underline{underlined}}. All hyper-parameters are summarized in Appendix Table~\ref{apptable:hyperparameters}.}
    \label{apptab:sparse}
\end{table*}

\begin{table*}[t!]
    \centering
    \footnotesize
        \begin{tabular}{@{}lcccccc|ccccc@{}}
        \hline 
        &  \multicolumn{5}{c}{\textbf{E2E}} & \multicolumn{5}{c}{\textbf{DART}} \\
        \textbf{Method} & \#Parm. & BLEU & NIST & MET & R-L & CIDEr & BLEU & MET & TER ↓ & Mover & BERT \\
        \hline
Finetune & 406M & \textbf{67.8} & \textbf{8.76} & \textbf{\underline{45.1}}& \textbf{\underline{69.5}} &\textbf{2.38} & 
        \textbf{\underline{46.1}} & \underline{\textbf{0.38}} & \textbf{0.46} & {0.53} & {0.95}  \\
        Prefixtune & 406M+\%0.1   &  67.3 & \textbf{\underline{8.66}} & 44.8& 68.6 & \textbf{\underline{2.34}} &
        45.9 & \textbf{0.39} & \underline{\textbf{0.45}} & {{0.53}} & {0.95} \\
        \hline
        HSoftSA &406M+\%0.1 & 66.2 & 8.57 &45.0  & 68.7 & 2.33 & \textbf{46.2} & \textbf{0.39} & \textbf{\underline{0.45}} & 0.53 & {0.95}  \\
        HierBlock+SoftSA & 406M+\%0.1& \textbf{68.0} & \textbf{{8.76}} & \textbf{{45.3}} & \textbf{69.6} & \textbf{{2.38}} & 46.0 & \textbf{0.39} & \textbf{0.46} & 0.53 & 0.95 \\
        \hline
    \end{tabular}
    \caption{\small \textbf{Data-to-Text}: \textbf{Hierarchical Sparse Attention} Models compared against Finetune and Prefixtune models. (\textbf{*}) All models use BART-Large as backbone and best models are \textbf{bolded}  and second best are also \textbf{\underline{underlined}}. For all the prefix-tuned models, the prefix-length is 5 for E2E and 10 for DART dataset. All hyper-parameters are summarized in Appendix Table~\ref{apptable:hyperparameters}.}
    \label{apptab:table2textsparsity}
\end{table*}



\subsection{Investigation of the Impact of Sparsity (Cont. from $\S$~\ref{doessparseattentionlhelp})}
\label{app:spectrum}

\paragraph{Spectrum Analysis:} 
We conduct spectrum analysis of the encoder attention matrix \textbf{A} zooming in on the prefix parameters to investigate if our sparse models do in fact learn sparse representations. A similar spectrum analysis has been used to prove the sparsity of the attention matrix in Linformer \citep{linformer}, a sparse transformer.  Our goal is to analyze the principal components of the subspace that captures the variation of the attention scores in prefix parameters. The eigenvalues capture the variation of the
attention scores distribution along different principal components. The higher the elbow in the spectrum graph, the less parameters are used and the model learns to represent the inputs with only the salient terms ignoring superfluous details. 

For our spectrum analysis, we compare the baseline prefix-tuning, which encodes a \textit{dense} attention matrix everywhere in the network (Dense PT) against one of our sparse prefix-tuned models with truncated attention matrix (Sparse PT), as we explained in $\S$~\ref{sparseattnpt}-(a), using top-\textit{p} sampling. Both models are a 12-layer stacked transformer (BART-LARGE) trained on XSum extreme summarization task. We apply
singular value decomposition into \textbf{A} across different layers and different heads of the model, and
plot the normalized cumulative singular value averaged over 1000 sentences. We compare the models' sparsity patterns at the top and at the lower layers separately as shown in Figure~\ref{fig:eigen}. 
The two figures exhibit a long-tail spectrum distribution across layers and heads. 
This implies that most of the
information of matrix \textbf{A} can be recovered from the first few largest singular values. 
We observe that the spectrum distribution in lower layers is more skewed than in
higher layers, meaning that, in lower layers, more information is concentrated in the largest singular
values and the rank of \textbf{A} is lower. With sparse attention at the lower layers and dense attention at the top layers, the prefix-tuned models can encode salient features controlling the generation.

\paragraph{Sparsity Analysis:} In Table~\ref{apptab:sparse} we show the ROUGE-1, ROUGE-2 and ROUGE-L scores of fine-tuned and prefix-tuned summarization models comparing dense and sparse attention impact. We observe that when sparsity is used on the prefix-parameters, the prefix-tuned models outperform dense counterparts. The performance improvements are more pronounced on shorter generation tasks such as XSUM but we still see improvements reacing up to 2.0 ROUGE-L score improvements on longer documents such as Wikihow. Similar performance patterns are observed in Table~\ref{apptab:table2textsparsity} on data-to-text generation tasks using E2E and DART.

\begin{table*}[t]
    \centering
    \footnotesize
    \begin{tabular}{l@{}c@{}c@{}c|@{}c@{}c@{}c|@{}c@{}c@{}c}
        \hline 
        Criteria & \textbf{Prefixtune} \ \ & \textbf{HierBlock} & & \ \ \textbf{Prefixtune} \ \ & \textbf{HSoftSA} & & \ \ \textbf{Prefixtune} \ \ & \textbf{HierBlock+SoftSA} & \\
         & wins & wins & same & wins & wins & same &  wins & wins & same \\
         \hline

          factuality &	36 &	36 &	78 & 25 & 45 & 80 & 39&	38&	73\\
          relevance &	31	& 29	&90 & 20 & \textbf{39} & 91&27&	33	&90\\
          gramaticality&	26	&28	&96 & 24 & 24 & 102 &28&	20&	102 \\
          coherence	&32&	32&	86 & 28 & \textbf{37} & 85&	32	&29&	89\\
          overall	&40&	38&	72 & 30 & \textbf{49} & 71&41	&38	&71\\
         \hline
         \hline
                 Criteria & \textbf{HierBlock} \ \ & \textbf{HSoftSA} & & \textbf{HierBlock} \ \ & \textbf{HierBlock+SoftSA} & & \textbf{HSoftSA} \ \ & \textbf{HierBlock+SoftSA} & \\
         & wins & wins & same & wins & wins & same &  wins & wins & same \\
         \hline
    
factuality &28	&\textbf{42}	&80	&28&	\textbf{42}&	80&	35&	41&	74 \\
relevance &25&	34&	91&	23&	35&	92&	31&	32&	87 \\
gramaticality&18&	\textbf{35}&	97&	26&	23&	101&	26&	19&	105\\
coherence	&18	&\textbf{44}	&88&	34&	30&	86&	35&	27&	88\\
 overall	&28	&\textbf{48}	&74	&32	&\textbf{44}	&74	&38&	38&	74\\
          \hline
                 \hline
                 Criteria & \textbf{Finetune} \ \ & \textbf{HierBlock} & & \textbf{Finetune} \ \ & \textbf{HierBlock+SoftSA} & & \ \ & & \\
         & wins & wins & same & wins & wins & same & & &\\
         \hline

factuality &\textbf{48} &	39&	63&	44&	42&	64 & & & \\
relevance &37 &     40&      73	&31	&\textbf{47}&	72 & & & \\
gramaticality & \textbf{38}&	24&	88&	\textbf{43}	&17	&90 & & & \\
coherence&\textbf{45}	&32	&73&	\textbf{46}&	30	&74 & & & \\
overall&45&	49	&56	&39&	\textbf{49}	&62 & & & \\
\hline
    \end{tabular}
    \vspace{-0.3cm}
    \caption{Head-to-Head comparison
of human evaluations on random subset of Wikihow
dataset.}
    \label{apptab:humandetails}
    \vspace{-0.3cm}
\end{table*}

\subsection{Investigation of the Impact of Sparsity on Hierarchically Blocked Prefixes (Cont. from $\S$~\ref{sparsityinhierblock})}
\label{app:sparsityinhierblock}
In Table~\ref{apptab:sparse} we showed ROUGE-L results of our hierarchical prefix blocking (HierBlock) model against hierarchical prefix blocking model with soft sparse attention (HierBlock+SoftSA). We observe improvements on performance on most summarization tasks including news summarization (XSum and CNN/DM), dialog summarization (SAMSum). 
We find that HierBlock+SoftSA models show much larger improvements on XSUM and Wikihow summarization,  from +0.5 to close to 2 ROUGE-L scores. 
On the structure to text generation tasks the sparsity on hierarchical blocking helps on some datasets (with E2E), though both HierBlock and HierBlock+SoftSA perform better than the baseline prefix-tuning models (see Table~\ref{apptab:table2textsparsity}).



\subsection{Automatic Evaluations (Cont. from $\S$~\ref{setup})}
\label{app:autoevals}
For model evaluations we use ROUGE-1/2/L using Python rouge-score 0.0.4 version licensed under the Apache 2.0 License. We use the default ROUGE script \href{https://github.com/GEM-benchmark/GEM-metrics/blob/bffcf6b6fc2ba240a9882aa2a86254cfb88bc517/gem_metrics/rouge.py}{rouge.py} from the GEM evaluation \href{https://gem-benchmark.com/resources}{shared task}.

\subsection{Human Evaluations (Cont. from $\S$~\ref{humanevals})}
\label{app:humanevals}

We perform human evaluations to establish that
our model’s ROUGE improvements are correlated
with human judgments. 
We compare the generations from four models: 
baseline prefix-tune (PT), 
Hierarchically Blocked PT (HierBlock/HB), Hierarchical Soft Sparse Attention PT (HSoftSA) and the ensemble of the blocked sparse model (HierBlock+SoftSA).
We use the following as evaluation criteria for generated summaries, which we include in the instructions for the annotators.  

\paragraph{Faithfulness:}Are the details in the summary fully consistent with the details in the source document?  The summary must not change any details from the source document. The summary also must not hallucinate any information that is not in the source document.

\paragraph{Relevance:}Does the summary capture the key points of the text? Are only the important aspects contained in the summary? Is there any extra/irrelevant information?

\paragraph{Grammaticality:} Considers the grammatical quality of each individual sentence in the summary. For each sentence, does it sound natural and grammatically correct? 

\paragraph{Coherence:}Does the summary form a cohesive, coherent whole? Is it well-written, well-structured and well-organized? Is it easy to follow? It should not be a heap of related information, but should build from sentence to sentence to a coherent body of information about a topic. 

\paragraph{Overall Quality:}Given the input context, is the summary satisfactory? Does the summary provide good quality information to the user? Is it helpful, informative and detailed enough given the information that’s contained in the text? Which summary of the two do you prefer best overall?

\paragraph{Annotator Details:}Human annotation was conducted by 9 professional raters (7 linguist raters, 1 linguist subject-matter-expert and 1 linguist) employed by an outsourcing company handling content moderation. All raters are monolingual native speakers of English; 6 have a minimum of high school degree or equivalent and 3 have a bachelor’s degree. Raters received compensation starting at \$18 per hour (which is close to 2.5 minimum wage in the state where the raters are located) and were also provided with Premium Differential as part of their contracts. Each rater conducted between 44 and 175 pairwise evaluations. Data collection protocol was reviewed by expert reviewers and received expedited approval as the data presented to the raters did not contain any sensitive or integrity-violating content. Participant consent was obtained as part of the non-disclosure agreement signed by each rater employee upon hire. All raters have also signed a sensitive content agreement that outlined the types of content they may encounter as part of their employment, associated potential risks and information and wellness resources provided by the outsourcing company to its employees.

\paragraph{Human Evaluation Procedure:} We randomly select 50 samples from the Wikihow test set and
ask 9 trained judges to evaluate them on the 5 criteria defined above. We perform head-to-head evaluation (more common in DUC style evaluations), where judges are shown the original document,
the ground truth summary and two model summaries in random order. The judges are then asked to compare two model summaries based on each of the five criteria. In each case, a judge either has the option to choose a model summary that ranks higher on a given criterion (i.e., respond by identifying the winning summary), or assert that both summaries are similar given the criterion and rate the comparison as "same". The evaluation of each pair of summaries across all 5 criteria takes on average between 5 and 10 minutes to complete. The raters were shown the data, as shown in Table~\ref{app:humanspreads}, to be rated in a spread sheet, where each line contained multiple columns in sequence: document, human written summary, model-A generated summary, model-B generation summary, and five additional columns indicating faithfulness, relevance, gramaticality, coherence, overall quality. The headers of the columns were clearly stated. The rates enter a/b/same in each corresponding cell when comparing summaries head-to-head based on each criteria.

\paragraph{Human Evaluation Results:} 
In Table~\ref{apptab:humandetails} we show head-to-head evaluation scores on all five metrics showing wins from each model as well as when both are selected as equal. Each sub-table compare a different model. 
Our Hierarhical Blocking (HierBlock) and  Hierarchical Soft Sparse Attention (HSoftSA) 
models beat prefix-tuning and HierBlock significantly ($p < .05$) beats most of our sparse models on all axes including factuality. In 

On a small data annotation, we also compare two of our best models 
HierBlock and HierBlock+SoftSA againts best finetuning model generations, which are shown in the same Table~\ref{apptab:humandetails}. We observe that in most cases both of our models are prefered as good as finetuning on all criteria, except on overall, the HierBlock summaries are ranked much higher than fine-tuning models.

			
			
			
			

\begin{table}[t]%
\centering
\begin{minipage}[c]{0.2\textwidth}%
\centering
     \begin{tikzpicture}[yscale=0.45, xscale=0.45] 
     \begin{axis}[
     font=\fontsize{15}{0}\selectfont,
     xmax=50,xmin=0,
     ymin= 36,ymax=41,
     xlabel=\emph{\% train samples},ylabel=\emph{Rouge-1},
     xtick={5,10,25,50},
     ytick={36,37,38,39,40,41},
     legend columns=2, 
     legend style={at={(0.5,1.15)},anchor=north},line width=1pt,
     ]


     \addplot [green!60!black,mark=square] coordinates{ (5,37.80) (10,38.47) (25,39.07) (50,39.95)};
     \addplot [olive,mark=x] coordinates{ (5,36.90) (10,37.76) (25,38.78) (50,39.30)};
     \addplot [blue,mark=*] coordinates{ (5,37.77) (10,38.47) (25,39.46) (50,40.07)};
     \addplot [red,mark=o] coordinates{ (5,38.14) (10,38.94) (25,39.62) (50,40.07)};
     
     \legend{\emph{Finetune},\emph{Prefix-tune},\emph{HierBlock},\emph{HierBlock+SA}}
     \end{axis} 
     \end{tikzpicture}
\vspace{-0.5cm}
\subcaption{\footnotesize Average Rouge-1}
\label{appfig:lowdata1}
\end{minipage}
\qquad
\begin{minipage}[c]{0.2\textwidth}%
\centering
     \begin{tikzpicture}[yscale=0.45, xscale=0.45] 
     \begin{axis}[
     font=\fontsize{15}{0}\selectfont,
     xmax=50,xmin=0,
     ymin= 14.3,ymax=17,
     xlabel=\emph{\% train samples},ylabel=\emph{Rouge-2},
     xtick={5,10,25,50},
     ytick={14,15,16,17},
     legend columns=2, 
     legend style={at={(0.5,1.15)},anchor=north},
     line width=1pt,
     ]


    \addplot [green!60!black,mark=square] coordinates{ (5,14.80) (10,15.28) (25,15.91) (50,16.72)};
     \addplot [olive,mark=x] coordinates{ (5,14.48) (10,15.07) (25,15.91) (50,16.38)};
      \addplot [blue,mark=*] coordinates{ (5,15.01) (10,15.48) (25,16.23) (50,16.74)};
     \addplot [red,mark=o] coordinates{ (5,15.25) (10,15.84) (25,16.43) (50,16.70)};
     \legend{\emph{Finetune},\emph{Prefix-tune},\emph{HierBlock},\emph{HierBlock+SA}}
     \end{axis} 
     \end{tikzpicture}
\vspace{-0.5cm}
\subcaption{\footnotesize Average Rouge-2}
\label{appfig:lowdata2}
\end{minipage} \\
\begin{minipage}[c]{0.2\textwidth}%
\centering
     \begin{tikzpicture}[yscale=0.45, xscale=0.45] 
     \begin{axis}[
     font=\fontsize{15}{0}\selectfont,
     xmax=50,xmin=0,
     ymin= 27.5,ymax=31.5,
     xlabel=\emph{\% train samples},ylabel=\emph{Rouge-L},
     xtick={5,10,25,50},
     ytick={27,28,29,30,31},
     legend columns=2, 
     legend style={at={(0.5,1.15)},anchor=north},line width=1pt,
     ]

     \addplot [green!60!black,mark=square] coordinates{ (5,28.45) (10,29.09) (25,30.53) (50,31.05)};
     \addplot [olive,mark=x] coordinates{ (5,27.87) (10,28.79) (25,29.78) (50,30.27)};
     \addplot [blue,mark=*] coordinates{ (5,28.54) (10,29.21) (25,30.09) (50,30.68)};
     \addplot [red,mark=o] coordinates{ (5,28.94) (10,29.71) (25,30.38) (50,30.60)};

          \legend{\emph{Finetune},\emph{Prefix-tune},\emph{HierBlock},\emph{HierBlock+SA}}
     \end{axis} 
     \end{tikzpicture}
\vspace{-0.5cm}
\subcaption{\footnotesize Average Rouge-L}
\label{appfig:lowdata3}
\end{minipage}

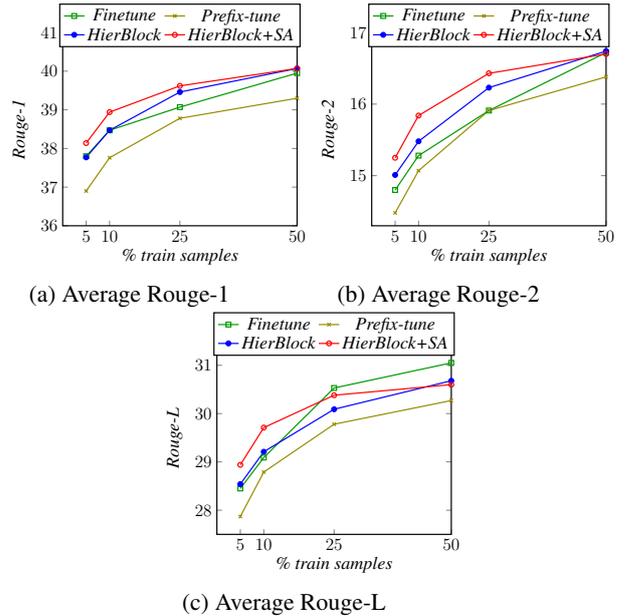
\captionof{figure}{\small Quantitative analysis on \textbf{low-resource settings}. The charts show \textbf{average} of ROUGE-1, ROUGE-2, ROUGE-L scores from models trained on two summarization tasks: XSUM and Wikihow. Our structurally biased parameter tuned HierBlock (\textcolor{blue}{blue}) and HierBlock+SA (\textcolor{red}{red}) consistently outperforms the baseline Finetuned (\textcolor{green!60!black}{green}) and Prefix-tuned models (\textcolor{olive}{olive}) when <50\% training data is used.}
\label{appfig:low}
\end{table}

\subsection{Low-data settings (Cont. from $\S$~\ref{lowdatasetting})}
\label{app:lowdatasetting}

In Figure~\ref{appfig:low}, we plot the ROUGE-1, ROUGE-2 and ROUGE-L scores averaging scores from two summarization tasks (XSUM and Wikihow). Our structured prefix parameter tuned models, HierBlock (\textcolor{blue}{blue}) and its sparse extension which uses sparse features, HierBlock+SA (\textcolor{red}{red})
outperforms Prefix-tuned models (\textcolor{olive}{olive}), while using the same number of parameters in low resources settings (when <50\% training samples are used). Both models outperform Finetuned models (\textcolor{green!60!black}{green}) on ROUGE-1 and ROUGE-2 metrics (Figure~\ref{appfig:low}-(a)\&(b)). While the HierBlock models show consistent performance, we conclude that on low-resource settings HierBlock-SA performance is more stable.

\clearpage

\end{document}